\documentclass[11pt]{article}
\usepackage{tcolorbox}
\usepackage[preprint]{acl}
\usepackage{xcolor}
\usepackage{enumitem}
\definecolor{keywordblue}{RGB}{0,0,180}
\definecolor{templategray}{gray}{0.9}
\usepackage{linguex}
\usepackage{microtype}
\usepackage{makecell}
\usepackage{caption}
\usepackage{bbm}
\usepackage{subcaption}
\usepackage{amsmath}
\usepackage{lipsum}
\usepackage{hyperref}
\usepackage{cleveref}
\usepackage{times}
\usepackage{latexsym}
\usepackage{tabularx}
\usepackage{amssymb} 
\usepackage[T1]{fontenc}
\usepackage[utf8]{inputenc}
\usepackage{multirow}
\usepackage{geometry}
\usepackage{microtype}
\usepackage{booktabs}
\usepackage{makecell}

\usepackage{inconsolata}
\usepackage{utfsym}
\usepackage{placeins}  %
\usepackage{dblfloatfix}
\usepackage{graphicx}

\newcommand{\ins}[1]{\colorbox{olive!20}{#1}}
\newcommand{\del}[1]{\colorbox{red!20}{#1}}
\newcommand{\shuf}[1]{\colorbox{teal!15}{#1}}

\title{Implicit Representations of Grammaticality in  Language Models}

\author{Yingshan Susan Wang\quad Linlu Qiu\quad Zhaofeng Wu\quad {Roger P. Levy} \quad  Yoon Kim  \vspace{1mm} \\ \vspace{1mm}
        Massachusetts Institute of Technology \\ \texttt{susanw26@mit.edu}   \vspace{-3mm}}

\begin{document}

\maketitle
\begin{abstract}
  \vspace{-1mm}
Grammaticality and likelihood are distinct notions in human language. Pretrained language models (LMs), which are probabilistic models of language fitted to maximize corpus likelihood, generate grammatically well-formed text and discriminate well between grammatical and ungrammatical sentences in tightly controlled minimal pairs. However, their string probabilities do not sharply discriminate between grammatical and ungrammatical sentences overall. But do LMs implicitly acquire a grammaticality distinction distinct from string probability? We explore this question through studying  internal representations of LMs, by training a linear probe on a dataset of grammatical and (synthetic) ungrammatical sentences obtained by applying perturbations to a naturalistic text corpus. We find that this simple \emph{grammaticality probe} generalizes to human-curated grammaticality judgment benchmarks and outperforms LM probability-based grammaticality judgments. When applied to semantic plausibility benchmarks, in which both members of a minimal pair are grammatical and differ in only plausibility, the probe however performs worse than string probability. The English-trained probe also exhibits nontrivial cross-lingual generalization, outperforming string probabilities on grammaticality benchmarks in numerous other languages. Additionally, probe scores correlate only weakly with string probabilities. These results collectively suggest that  LMs acquire to some extent an implicit grammaticality distinction within their hidden layers.\footnote{The code for this project is available at \url{https://github.com/SusanWYS/grammaticality\_probe}.}

\end{abstract}

  \vspace{-1mm}
\section{Introduction}
\vspace{-1mm}
\begin{figure}[!t]
  \centering
\includegraphics[width=1\columnwidth]{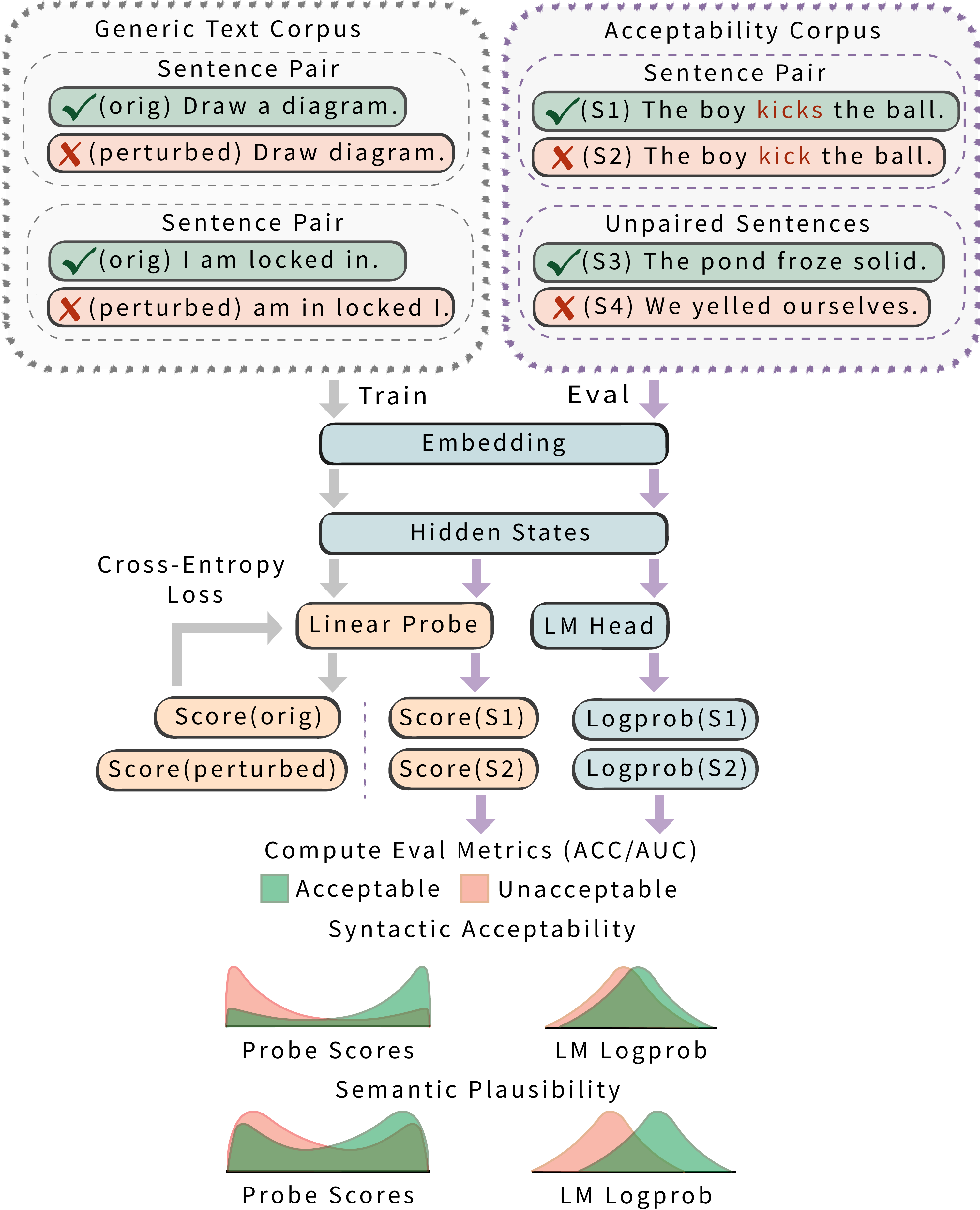}
  \vspace{-2mm}
  \caption{The grammaticality probe is \textbf{a linear classifier over LM hidden states}, trained on a synthetic dataset created from perturbing  sentences from a generic text corpus. For evaluation, we measure how well probe scores and LM logprob discriminate between grammatical and ungrammatical sentences on acceptability-judgment corpora.  As a baseline, we also evaluate the grammaticality probe on semantic plausibility datasets. \textbf{The probe is more discriminative for syntactic acceptability than LM logprobs, whereas LM logprobs are more sensitive to semantic plausibility.}}
  \label{fig:main_diagram}
  \vspace{-2mm}
\end{figure}
In human language, a  sentence's \emph{grammaticality}---whether it conforms to the rules of language---is distinct from its 
\emph{likelihood}---whether it is likely to occur in naturalistic text. Consider the following pair of examples from \citet{levy-etal:2025-science-of-language} (itself adapted from \citet{BOCK199145}):

\ex. {*The key to the cabinets are on the table.} \label{ex:key-cabinets-are}

\ex. {The key to the cabinets was so thoroughly rusted that it was impossible to fit into the keyhole.} \label{ex:key-cabinets-was}

The first sentence violates the rule of English that the number of the verb \emph{are} should match that of the singular head noun of the subject \emph{key}, and is therefore ungrammatical. However, this type of error is common when a plural noun (\emph{cabinets}) intervenes before the verb, so the sentence is almost certainly more likely to be used than the second sentence, which is grammatical but expresses a complex and unusual meaning. Insofar as this \emph{grammaticality--likelihood distinction}---a  consequence of the broader \emph{competence--performance distinction}---is present in humans, the extent to which LMs are able to capture this distinction bears directly on their viability as candidate computational models of human language processing and acquisition---a matter of considerable ongoing debate in linguistics and cognitive science  \citep[][\emph{i.a.}]{piantadosi2023modern,katzir2023large,milliere2024language,fox2024large,futrell2025linguistics}.

LMs are at their core statistical systems that assign probabilities to strings, and thus naturally capture a notion of likelihood---they are literally trained to maximize the likelihood of the training data. However, their training data inevitably contain noise arising from (amongst others) \emph{performance} factors, and sentences such as (1) occur frequently in naturalistic text. Thus, LMs do not (nor are they designed to) categorically assign lower probabilities to ungrammatical sentences than grammatical ones; indeed, \citet{levy-etal:2025-science-of-language} note that GPT-2 assigns 146 trillion times more probability to (2) than to (1), and prior studies  find that LM probabilities are generally  poor at distinguishing grammatical from ungrammatical sentences \citep{leivada2024evaluating,leivada2024reply,leivada2025large}. \citet{hu-etal:2025-what-can-string-probability} further show that string probabilities alone are theoretically inappropriate quantities for extracting grammaticality judgments under a simple generative model of language. 
Nevertheless, LM probabilities have been shown to be  effective at adjudicating grammaticality across \emph{pairs} of sentences that differ from one another minimally \citep[][\emph{i.a.}]{Linzen2018,futrell-etal-2019-neural,warstadt2020blimp,hu-etal-2020-systematic}, indicating that while the grammaticality--likelihood distinction is not explicitly baked into LMs, they  acquire behavioral generalizations that indicate some knowledge of a grammar.
But as \citet{leivada2024evaluating} note, this type of grammaticality assessment based on minimal pairs is different from the standard by which we assess  grammaticality judgments in humans. 

However, while string probabilities from LMs are not designed to (only) encode grammaticality, in neural LMs these string probabilities are the {result} of  internal computations performed by the model. In the present work we thus explore the hypothesis that LMs, through large-scale training, learn to \emph{implicitly} encode grammaticality judgments within their {hidden representations}.\footnote{\emph{Metalinguistic} judgments, wherein an LM is explicitly asked to answer whether a sentence is grammatical or not, provides another perspective on the grammaticality--likelihood distinction in LMs. While this approach ostensibly mirrors how we might extract grammaticality judgments from adult native speakers, \citet{hu2023} observe that metalinguistic judgments can underestimate a model's linguistic capability. Moreover, this type of assessment assumes that an LM is able to perform basic question-answering out of the box, which may not be the case for many LMs of interest (e.g., smaller LMs that have not been instruction-tuned). Hence, this work does not use it as a major baseline.} A standard approach to testing such hypotheses is through supervised probes, wherein a simple model (e.g., a linear classifier) is trained to predict some phenomena of interest \citep{alain2018, Belinkov2022}. In the case of grammaticality judgments, this might involve training a probe on acceptability judgment datasets on human-curated datasets such as CoLA \citep{warstadt2019} or BLiMP \citep{warstadt2020blimp}. However, such datasets are often  small and costly to collect, and thus probes trained on them could overfit; ideally, such datasets  should be used as held-out sets for evaluation.

\begin{table*}[t]
\centering
\small
\setlength{\tabcolsep}{5pt}
\renewcommand{\arraystretch}{1.15}

\begingroup
\setlength{\fboxsep}{1pt}

\begin{tabularx}{\textwidth}{@{}>{\raggedright\arraybackslash}m{0.10\textwidth} X X@{}}
\toprule
\textbf{Perturbation} & \textbf{Original} & \textbf{Perturbed} \\
\midrule

\multirow{3}{*}{\parbox[c]{0.10\textwidth}{\raggedright\textbf{\ins{Insertion}}}} &
Analysts agreed. &
\ins{Democratization} Analysts \ins{constructed} agreed. \\

&
``They had to do it.'' &
``They had to \ins{Krenz} do it.'' \\

&
She bursts into tears and walks away. &
She bursts into tears \ins{fringe} and walks away. \\

\midrule

\multirow{3}{*}{\parbox[c]{0.10\textwidth}{\raggedright\textbf{\del{Deletion}}}} &
What \del{is} his \del{usual} \del{food}, Nelly? &
What his , Nelly? \\

&
\del{Be} \del{a} good lad; \del{and} I’ll \del{do} \del{for} you. &
Good lad; I’ll you. \\

&
They allowed \del{this} \del{country} to be credible. &
They allowed to be credible. \\

\midrule

\multirow{3}{*}{\parbox[c]{0.10\textwidth}{\raggedright\textbf{\shuf{Local shuffle}}}} &
It was \shuf{jolly of you to make} up your mind to come. &
It was \shuf{of you to make jolly} up your mind to come. \\

&
Now \shuf{we shall have some discussion}, we’ll see to that. &
Now \shuf{have discussion shall we some}, we’ll see to that. \\

&
\shuf{Have you been reading Spencer}? &
\shuf{you reading Spencer Have been}? \\

\bottomrule
\end{tabularx}

\endgroup
\caption{Original sentences and their perturbed counterparts for three perturbation types.}
\vspace{-3mm}
\label{tab:perturb_examples}
\end{table*}

We propose an approach for learning a grammaticality probe without relying on human-annotated data. We create a synthetic dataset of ``good'' and ``bad'' sentences by applying noise to sentences in a generic text corpus, where we randomly insert, delete, or shuffle tokens to create (mostly) ungrammatical examples. We find that a linear probe trained on this dataset can outperform probability-based grammaticality judgments in both minimal-pair- and non-minimal-pair-based grammaticality judgment benchmarks. The probe surprisingly exhibits cross-lingual generalization, i.e., a probe trained on English data can be used for grammaticality judgment on other languages, and outperforms probability-based judgments. The outperformance of probes over probabilities is flipped when distinguishing between semantically plausible/implausible sentences, and the probe scores are found to weakly correlate with string probabilities.

In arguing against LMs as theories of human linguistic cognition, \citet{katzir2023large} state that the grammaticality--likelihood distinction is ``entirely foreign'' in LMs. Our results however suggest that this may not necessarily be the case, and that the internal representations of LMs may encode a notion of grammaticality that is distinct from string probabilities.\footnote{Of course, there still remain many unaddressed arguments against LMs as  models of human language processing.}

\section{A Grammaticality Judgment Probe}

To explore whether grammaticality is implicitly encoded in the hidden states of LMs, we train a probe on a dataset of (mostly) grammatical and (mostly) ungrammatical sentences obtained from applying noise to a generic text corpus. 
\subsection{Generation of Contrastive Training Set} The training set of the probe comprises sentences with synthetically-derived binary acceptability labels. We sample 50,000 sentences from the Penn Treebank \citep{marcus1994} and Project Gutenberg \citep{gutenbergdpo} to form a generic text dataset as grammatical examples. To generate negative ungrammatical examples, we use three simple perturbation functions based on prior work \citep{cotterell2018,cao-etal-2020-unsupervised, mitchell-bowers2020,kallini2024}: 
\begin{itemize}
  \setlength{\itemsep}{2pt}
  \setlength{\parskip}{2pt}
  \setlength{\parsep}{2pt}
    \item \textbf{Insertion: }inserts 1–5 random  tokens from the entire corpus at random position in the sentence.
    \item 
    \textbf{Deletion: }randomly removes 1–5 text tokens.
    \item \textbf{Local shuffle: }randomly permutes a contiguous 5-token window.
\end{itemize}
See \hyperref[tab:perturb_examples]{Table~\ref*{tab:perturb_examples}} for examples of perturbations and \S \ref{Perturb} for generation details. We perturb every sentence with a (uniformly) randomly chosen noising function. Sentences that cannot be perturbed are filtered out (e.g., three-word sentences assigned to local shuffle). We label originals as grammatical and perturbed sentences as ungrammatical.  80\% of the data is used as train set and 20\% as dev set for hyperparameter tuning.

The data collection is not perfect: some original sentences may be ungrammatical, and our perturbations may introduce semantic implausibility instead of syntactic errors. To validate our data generation pipeline, we use an LLM to judge the sentence acceptability of 5,000 randomly sampled sentence pairs, confirming that 93.72\% of the perturbed subset is indeed ungrammatical (\S \ref{Perturb}). Furthermore, natural ungrammatical sentences are rarely simple insertion, deletion, or shuffle variants of grammatical text. Demonstrating that a synthetically-trained probe can generalize to real grammaticality benchmarks would thus be a significant result.

\subsection{Linear Probe}
Given an LM $\mathcal{M}$ and a sentence-label pair $(s_i,y_i)$ with sentence $s_i \in \Sigma^{\ast}$ and label $y_i \in \{0, 1\}$, let $h_i := \mathcal{M}(s_i)$ be the hidden states taken at the last token (generally a punctuation token) from an LM $\mathcal{M}$ applied on sentence $s_i$. Here $h_i$ is either the hidden states of a single layer or all layers concatenated, a design choice we will specify in the experimental setup. 
We train a logistic classifier with $\ell_2$ regularization: 

\begin{equation}
\label{eq:probe_objective}-\frac{1}{N} \sum_{i=1}^N \log p(y_i | s_i) +
\alpha\|w\|_2^2.
\end{equation}
where $\log p(y_i | s_i)$ is $ y_i\log \left( \sigma(w\cdot h_i +b)\right) + (1-y_i)\log(1-\sigma(w\cdot h_i + b))$ and  $\alpha$ is tuned as a hyperparameter on the dev set. We also experiment with $\ell_1$ regularization in our later experiments.

\section{Experimental Setup}
\subsection{Models} Our experiments use the following open-weight base models: OLMo-2-7B \citep{olmo2} and OLMo-3-7B \citep{olmo3}, Llama-3.2-1B and Llama-3.1-8B \cite{llama3}, and Gemma-2-2B and Gemma-2-9B \cite{gemma2}. We do not evaluate on instruction-tuned models because instruction tuning alters the next-token distribution in task- and alignment-specific ways, which would confound our analysis.

\subsection{Evaluation Metrics} Let $f(s)$ be a score for sentence $s$ obtained from the probe or (as our baseline) the string probabilities from an LM.\footnote{For string probabilities we use the  length-normalized cumulative logprobs, $\frac{1}{T} \sum_{t = 1}^T\log p_\theta(w_t | w_{<t})$, which was found to perform better than  cumulative probabilities.} We consider two metrics, depending on whether the benchmark is based on minimal pairs or not:

\paragraph{Minimal pairs.} Given a grammaticality judgment benchmark such as BLiMP which provides pairs of sentences $(s_i, s_i')$ that differ from one another minimally, we compute the accuracy (\textbf{ACC}):
\[ACC=\frac{1}{N}\sum_{i=1}^N \mathbbm{1}(f(s_i) > f(s_i')).\] where $s_i$ is the grammatical one.

\paragraph{Standalone acceptability.} On benchmarks such as CoLA which just provide a sentence and its label $(s_i, y_i)$, using pure accuracy would require setting a threshold on $f(s)$ for deciding when something is considered grammatical or not. For a binary classifier, a natural threshold would be 0.5, but the threshold for LM probabilities can be challenging to determine. Accuracy is also not always appropriate in cases where the positive rate differs from 50\%, which is indeed the case in some of our benchmarks.   We thus instead compute  area under the ROC curve (\textbf{AUC}):
\begin{multline*}
    AUC = \frac{1}{n_+ n_-}\sum_{s\in P}\sum_{s'\in N}\big( \mathbbm{1}\left(f(s) > f(s')\right)\\
 + \frac{1}{2}\mathbbm{1}\left(f(s) = f(s')\right)\big).
\end{multline*}
Here $P=\{s_i:y_i=1\}$ and  $N=\{s_i:y_i=0\}$, and we further have $n_+ = |P|$ and $n_- = |N|$.
Intuitively, AUC quantifies how consistently a random acceptable $s$ sentence receives a higher score than a random unacceptable sentence $s'$, and provides a more granular window into a classifier's performance than pure accuracy. We compute AUC on minimal-pair benchmarks as well.

\subsection{Evaluation Benchmarks} Our primary experiments test on three acceptability datasets in English: BLiMP, CoLA, and SyntaxGym \citep{warstadt2020blimp, warstadt2019, gauthier2020}. For cross-lingual generalization, we also test on six multilingual acceptability sets: Swedish (ScaLA (sv)) \citep{nielsen2023}, Dutch (BLiMP-NL) \citep{suijkerbuijk2025}, Italian (ItaCoLA) \citep{Trotta2021}, Russian (RuCoLA) \citep{mikhailov2022}, Japanese (JCoLA) \citep{someya2024}, and Chinese (SLING) \citep{song2022}. We compute the evaluation metric ACC on minimal-pair data only: BLiMP, BLiMP-NL, and SLING. All other datasets are single sentences with binary acceptability labels, for which we only calculate AUC. Dataset statistics are listed in \hyperref[tab:syntactic_sets]{Table~\ref*{tab:syntactic_sets}} and examples from the different benchmarks are shown in \hyperref[tab:dataset-examples]{Table~\ref*{tab:dataset-examples}} of the appendix.

In addition to benchmarks which primarily test for grammaticality, we also evaluate our probe on three semantic plausibility benchmarks collectively referred to as ``plausibility sets'' \citep{eventknowledge,fedorenko2020,vassallo2018,Ivanova2021}: see \hyperref[tab:plausibility_sets]{Table~\ref*{tab:plausibility_sets}} of the appendix for details.  LM probabilities entangle  grammaticality with likelihood, and thus their performance on such benchmarks have been shown to be quite high \citep{leivada2025large,hu-etal:2025-what-can-string-probability}. A grammaticality probe that is mostly sensitive to syntax should ideally have low performance on these plausibility benchmarks.
\section{Implicit Representations of Grammaticality in Language Models}\label{main}

Can probes trained on unsupervised, synthetic sentence pairs generalize to out-of-distribution human-labeled linguistic acceptability benchmarks? For our main results, we train grammaticality probes layer-by-layer and select the best layers as well as the regularization parameters $\alpha$ based on the synthetic dev set~(see \S\ref{l2} for training details).\footnote{We find that the middle layers generally perform best. The performance by layer for all models is shown in \hyperref[fig:by_layer]{Figure~\ref*{fig:by_layer}} of the appendix. \hyperref[fig:dev_final_metric]{Figure~\ref*{fig:dev_final_metric}} compares the performance of selected $\ell_2$-probes on the held-out dev set against LM logprob.}

As shown in \hyperref[fig:syntactic_test]{Figure~\ref*{fig:syntactic_test}}, our probes surpass the performance of probability-based method in both the minimal-pair setting (BLiMP) and the unpaired setting (BLiMP, CoLA, SyntaxGym), suggesting that grammaticality to an extent is implicitly captured by the hidden states of LMs. See \hyperref[tab:blimp_results]{Table~\ref*{tab:blimp_results}} in the appendix for results broken down by different grammaticality judgment categories.\footnote{As a supervised baseline we also train probes on BLiMP and evaluate them on CoLA, SyntaxGym, and our synthetic data. The probe trained on specific violations (agreement errors, island effects, etc.) successfully detects crude perturbations (insertions, deletions, shuffles). See \S\ref{supervised_probe}.}

\subsection{Grammaticality or Semantic Plausibility?}
\begin{figure}[t!]
  \centering
  \begin{subfigure}{\linewidth}
    \centering
    \includegraphics[width=\linewidth]{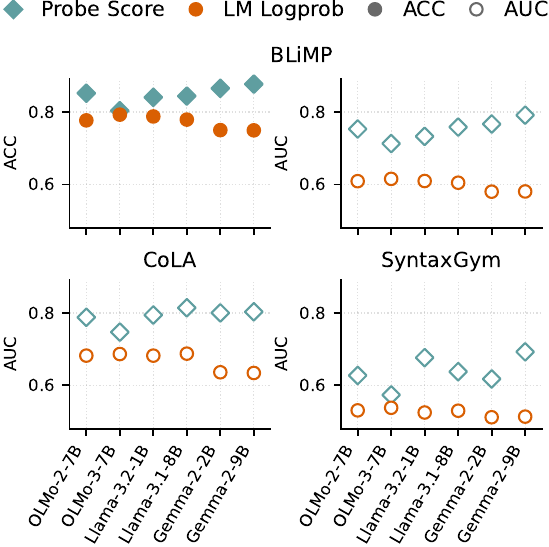}
    \caption{\textbf{Probe scores outperform LM string logprob baselines on grammaticality acceptability judgment datasets} (ACC and AUC on BLiMP; AUC on all datasets).}
    \label{fig:syntactic_test}
  \end{subfigure}

  \vspace{10pt}

  \begin{subfigure}{\linewidth}
    \centering
    \includegraphics[width=\linewidth]{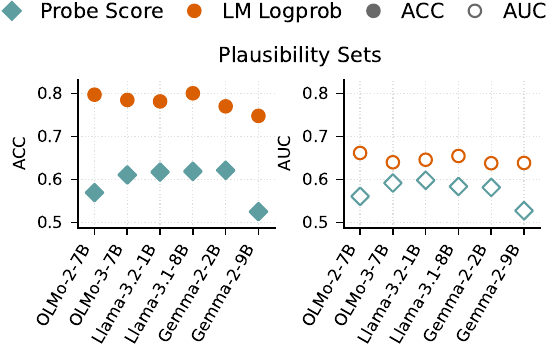}
    \caption{Probe scores underperform LM logprob on semantic plausibility  benchmarks, suggesting that \textbf{probes are more selectively tuned to syntactic acceptability}.}
    \label{fig:plausibility_test}
  \end{subfigure}
  \vspace{-5mm}
  \caption{Comparing probe scores and string logprob baselines on grammaticality and semantic plausibility benchmarks. We also compute their 95\% confidence intervals in the appendix (\S\ref{CI}).}
  \label{fig:probe_vs_logprob}
  \vspace{-5mm}
\end{figure}

One reason why string probabilities are inappropriate for adjudicating grammaticality is that they entangle grammaticality with other aspects of language such as meaning ~\citep{hu-etal:2025-what-can-string-probability}. Such entanglement  means that string probabilities can   distinguish between  semantically plausible/implausible pairs quite well, often  better than  grammatical/ungrammatical pairs ~\citep{leivada2025large,kauf2024}. Is our probe also similarly entangling grammaticality with plausibility? The results of our probe on the semantic plausibility benchmarks, shown in \hyperref[fig:plausibility_test]{Figure~\ref*{fig:plausibility_test}}, suggest otherwise. We find that our probe's performance on semantic plausibility sets is substantially lower than string probabilities, suggesting that  the LMs have potentially learned grammaticality representations that are distinct from semantic plausibility.

\subsection{Cross-lingual Generalization}
Given that LMs are implicitly capturing grammaticality within the hidden layers, are these features specialized to English, or are they language-agnostic? We next test whether our probes, trained on English data, zero-shot generalize to other languages. We study languages spanning multiple families, ranging from those typologically closer to English (Swedish and Dutch) to more distantly related ones (Italian and Russian), and finally to typologically distant languages (Japanese and Chinese). The results for Llama and Gemma models are listed in \hyperref[tab:multilingual]{Table~\ref*{tab:multilingual}}.\footnote{We do not test OLMo models on multilingual sets since they have limited multilingual capabilities.} Overall, the English-trained probes exhibit nontrivial cross-lingual generalization, and generally outperforms string probability-based grammaticality judgments across languages.

We further test the probes trained on non-English data and find less impressive transfer results, which we report in \S\ref{non_english_transfer}. This observation could potentially be explained by the disproportionate scale of English relative to other languages in the pretraining data, and confirms previous findings of an English-dominant representation space in pretrained models \cite{Llama_English}.

\begin{table*}[tbp]
\centering
\small %
\setlength{\tabcolsep}{4pt} %

\begin{tabular}{@{}llcccccccc@{}}
\toprule
& &
\multicolumn{1}{c}{\textbf{Swedish}} &
\multicolumn{2}{c}{\textbf{Dutch}} &
\multicolumn{1}{c}{\textbf{Italian}} &
\multicolumn{1}{c}{\textbf{Russian}} &
\multicolumn{1}{c}{\textbf{Japanese}} &
\multicolumn{2}{c}{\textbf{Chinese}} \\
\cmidrule(lr){3-3} \cmidrule(lr){4-5} \cmidrule(lr){6-6} \cmidrule(lr){7-7} \cmidrule(lr){8-8} \cmidrule(lr){9-10}

\multirow{2}{*}{\textbf{Model}} &
\multirow{2}{*}{\textbf{Method}} &
\multicolumn{1}{c}{\textbf{ScaLA (sv)}} &
\multicolumn{2}{c}{\textbf{BLiMP-NL}} &
\multicolumn{1}{c}{\textbf{ItaCoLA}} &
\multicolumn{1}{c}{\textbf{RuCoLA}} &
\multicolumn{1}{c}{\textbf{JCoLA}} &
\multicolumn{2}{c}{\textbf{SLING}} \\
\cmidrule(lr){3-3} \cmidrule(lr){4-5} \cmidrule(lr){6-6} \cmidrule(lr){7-7} \cmidrule(lr){8-8} \cmidrule(lr){9-10}

 & & \textbf{AUC} & \textbf{AUC} & \textbf{ACC} &
   \textbf{AUC} & \textbf{AUC} & \textbf{AUC} &
   \textbf{AUC} & \textbf{ACC} \\
\midrule

\multirow{2}{*}{Llama-3.2-1B}
 & LM Logprob & 0.62 & 0.58 & \textbf{0.75} & 0.58 & 0.44 & 0.57 & 0.55 & 0.59 \\
 & Probe Score  & \textbf{0.68} & \textbf{0.59} & 0.69 & \textbf{0.59} & \textbf{0.59} & 0.57 & \textbf{0.57} & \textbf{0.63} \\
\midrule

\multirow{2}{*}{Llama-3.1-8B}
 & LM Logprob & 0.65 & 0.62 & \textbf{0.84} & 0.61 & 0.46 & 0.59 & 0.57 & 0.65 \\
 & Probe Score & \textbf{0.74} & \textbf{0.65} & 0.77 & \textbf{0.65} & \textbf{0.58} & \textbf{0.63} & \textbf{0.60} & \textbf{0.69} \\
\midrule

\multirow{2}{*}{Gemma-2-2B}
 & LM Logprob & 0.64 & 0.61 & \textbf{0.82} & 0.55 & 0.46 & 0.59 & 0.57 & 0.61 \\
 & Probe Score & \textbf{0.69} & \textbf{0.62} & 0.73 & \textbf{0.62} & \textbf{0.61} & \textbf{0.61} & \textbf{0.63} & \textbf{0.73} \\
\midrule

\multirow{2}{*}{Gemma-2-9B}
 & LM Logprob & 0.66 & 0.63 & \textbf{0.86} & 0.56 & 0.47 & 0.59 & 0.58 & 0.62 \\
 & Probe Score & \textbf{0.83} & \textbf{0.73} & 0.84 & \textbf{0.70} & \textbf{0.65} & \textbf{0.70} & \textbf{0.67} & \textbf{0.75} \\
\bottomrule
\end{tabular}
\caption{Evaluation results of grammaticality probes and LM logprob on multilingual acceptability benchmarks, where the grammaticality probes are trained on synthetic English training data and tested zero-shot on the benchmark in each language. \textbf{The probes generally match or surpass the performance of logprob on syntactic acceptability judgments across languages.} Bold numbers indicate better performance. }
\label{tab:multilingual}
\end{table*}

\subsection{Localization to Select Neurons}
We have observed that grammaticality judgments can be probed out from LM representations and can generalize across languages. Is this phenomenon represented in a distributed manner, or can it be localized to a small number of neurons? We train an $\ell_1$-regularized linear probe (i.e., LASSO \citep{tibshirani1996regression}) on the concatenation of the hidden states from all layers, where we  vary the regularization strength  until we reach the desired percentage of nonzero weights. We target nonzero rates of \{0.01\%, 0.05\%, 0.1\%, 0.5\%\}.  We then retrain  $\ell_2$-regularized probes on top of the LASSO-selected neurons (see \S\ref{l1} for details). As a baseline, we also train $\ell_2$ probes on random subsets of neurons of the same size across 30 seeds and report the average performance. 

\begin{figure}[!t]
  \includegraphics[width=\columnwidth]{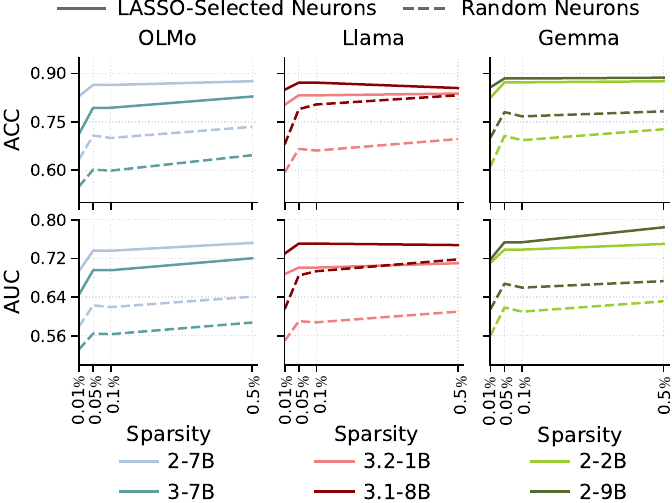}
  \caption{The evaluation results of LASSO probes on BLiMP. \textbf{Grammaticality signals are captured by very small (even random) subsets of neurons.} The performance gap between LASSO-selected and random neurons increases at higher sparsity.}
  \label{fig:lasso}
\end{figure}

The nontrivial performance of the random-neuron probes suggests that grammaticality signals are distributed across most layers (consistent with the by-layer $\ell_2$ probe in \hyperref[fig:by_layer]{Figure~\ref*{fig:by_layer}}). Nevertheless, a smaller subset of neurons carries a disproportionately rich share of this signal. We visualize the results on BLiMP here, and the test results for the remaining benchmarks are shown in \hyperref[fig:other_english_l1]{Figure~\ref{fig:other_english_l1}} and \hyperref[fig:multilingual_l1]{Figure~\ref{fig:multilingual_l1}}. As shown in \hyperref[fig:lasso]{Figure~\ref*{fig:lasso}}, even with 0.01\% of the neurons (which corresponds to about 10 neurons; see neuron counts in \hyperref[fig:l1_hist]{Figure~\ref{fig:l1_hist}}) the probes work well, suggesting that grammaticality can potentially be localized to a handful of neurons.

\section{Revisiting  Grammaticality vs.\ String Probabilities}
In the previous section (\S \ref{main}), we showed that grammaticality probes outperform string probabilities at predicting grammaticality. Are these scores distinct from string probabilities? If so, then this may provide the basis for a grammaticality--likelihood distinction in LMs.

\begin{figure}[!t]  \includegraphics[width=\columnwidth]{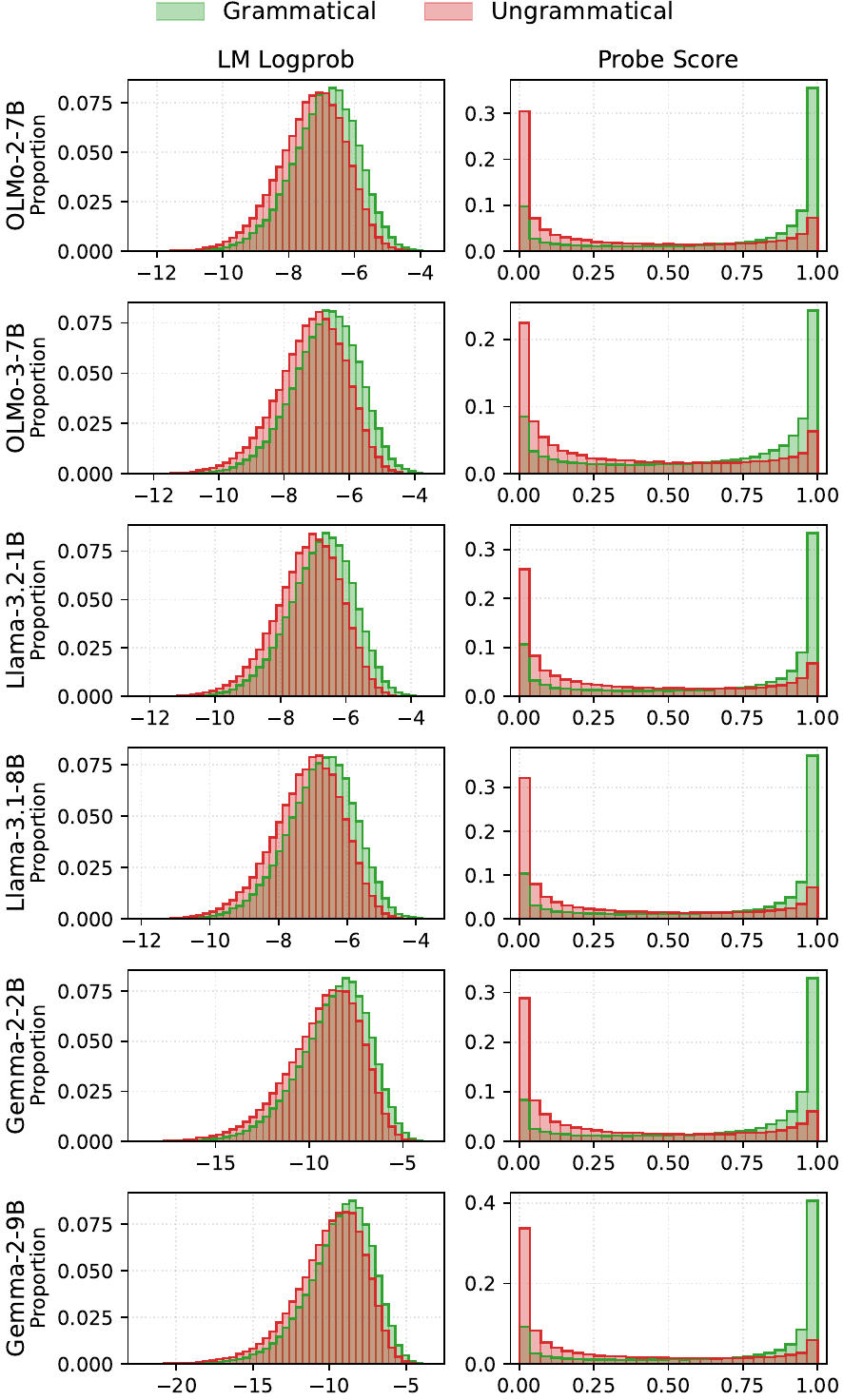}
  \caption{Distributions of LM logprob and probe scores on BLiMP. \textbf{Probe scores better separate grammatical and ungrammatical sentences than LM logprob.}}    \label{fig:dist_blimp}
\end{figure}

As a first step, we visualize the distribution of LM logprob and probe scores for BLiMP in \hyperref[fig:dist_blimp]{Figure~\ref*{fig:dist_blimp}}, where we show the probe scores in probability space for easier visualiation. The distributions for the other two English benchmarks are shown in \hyperref[fig:dist_cola]{Figure~\ref*{fig:dist_cola}} and \hyperref[fig:dist_syngym]{Figure~\ref*{fig:dist_syngym}}. As noted in previous work \citep{leivada2025large}, LM logprob fails to reliably distinguish grammatical from ungrammatical sentences; the grammaticality probe scores however achieve relatively good separation. We then compute the Spearman's correlation between LM logprob and log probe scores (\hyperref[tab:cor]{Table~\ref*{tab:cor}}). We observe only a moderate correlation, which further suggests that the probe scores are capturing distinct information and are not merely a reparameterization of the model's output probabilities.\footnote{We also compute the Pearson's correlation between the length normalized logprob and log probe scores, whose results are in \hyperref[tab:PearsonR]{Table~\ref*{tab:PearsonR}}, confirming that the two variables are not strongly linearly correlated.}

The observations above motivate a natural question: do string probabilities contain useful extra signal for acceptability beyond the probes, or is the logprob information already implicit in the representations the probe uses? 

\subsection{String Probabilities as Additional Features in the Probe}

\begin{figure}[t]  \includegraphics[width=\linewidth]{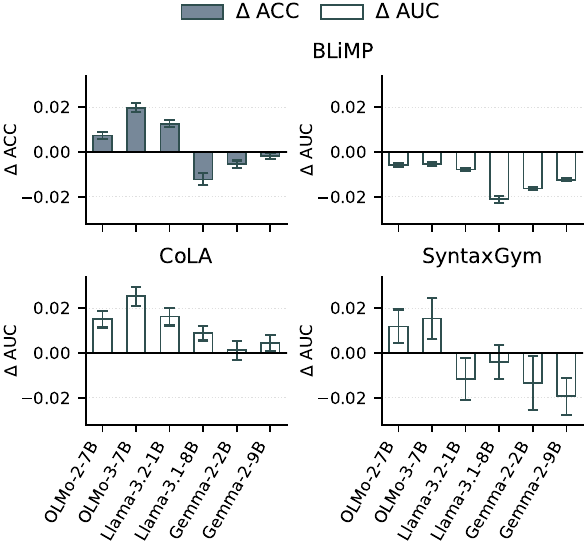}
  \caption{Performance difference ($\Delta$) between $\ell_2$ probes augmented with logprob as an input feature and the baseline $\ell_2$ probes. Error bars denote 95\% confidence intervals. \textbf{The string probabilities add little complementary signals to original probe scores.}}

  \label{fig:delta}
  \end{figure}

We train a probe with the same setup as in the previous section  but with logprob added as an extra feature (see \S\ref{probe_plus_prob} for details). \hyperref[fig:delta]{Figure~\ref*{fig:delta}} reports the performance change when augmenting the $\ell_2$ probe with length-normalized logprob ($\Delta$ = augmented $-$ baseline). Across BLiMP, CoLA, and SyntaxGym. We do not observe a consistent improvement across model families or datasets. Overall, these results suggest that probabilities provide little complementary signal once the probe already has access to the selected-layer hidden states.

\subsection{Probing for String Probabilities} 
The results of probes trained with logprobs as an extra predictor suggest that likelihood-related information that is relevant for grammaticality may already be encoded in the representations used by the probe. We now move on to a more direct test: can logprobs be recovered from the hidden states?

\begin{table}[tbp]
    \centering
    \small

    \begin{tabular}{lccc}
        \toprule
        & \multicolumn{3}{c}{\textbf{Corr.\,(logprob, probe score)}} \\
        \cmidrule(lr){2-4}
        \textbf{Models} & \textbf{BLiMP} & \textbf{CoLA} & \textbf{SyntaxGym} \\
        \midrule
        Llama-3.2-1B & 0.31 & 0.40 & 0.089 \\
        Llama-3.1-8B & 0.27 & 0.43 & 0.28 \\
        \midrule
        Gemma-2-2B   & 0.27 & 0.35 & 0.24 \\
        Gemma-2-9B   & 0.23 & 0.37 & 0.28 \\
        \midrule
        OLMo-2-7B    & 0.23 & 0.34 & 0.30 \\
        OLMo-3-7B    & 0.30 & 0.47 & 0.23 \\
        \bottomrule
    \end{tabular}
    \caption{Spearman's correlation evaluates the extent to which two variables exhibit a monotonic relationship. \textbf{Correlations between logprob and probe scores are only moderate, suggesting that a higher LM logprob does not always correspond to a higher probe score.}}
    \label{tab:cor}
\end{table}

  \begin{table}[h]
    \centering
    \small
    
    \begin{tabular}{lccc}
        \toprule
        & \multicolumn{2}{c}{\textbf{Hidden states $\rightarrow$ logprob ($R^2$)}} \\
        \cmidrule(lr){2-3}
        \textbf{Models} & \textbf{Per Token} & \textbf{Last Token Only}\\
        \midrule
        Llama-3.2-1B & 0.55 & 0.59 \\
        Llama-3.1-8B & 0.67  & 0.65 \\
        \midrule
        Gemma-2-2B   & 0.73 & 0.68 \\
        Gemma-2-9B   & 0.56  & 0.66 \\
        \midrule
        OLMo-2-7B    & 0.59  & 0.58 \\
        OLMo-3-7B    & 0.69  & 0.51 \\
        \bottomrule
    \end{tabular}

    \caption{Ridge regression $R^2$ for predicting length-normalized cumulative logprob from token-level hidden states. The per-token setups consider all time steps, while the last-token-only setup only uses final tokens. \bf{The moderately strong $R^2$ confirms that substantial information of prefix logprob can be recovered from token-level hidden states.}}
    \label{tab:ridge}
\end{table}

We investigate this under two setups: a \emph{per-token} approach utilizing the hidden state at each time step, and a \emph{last-token-only} approach using the last-token hidden state. Both setups predict the length-normalized cumulative log-probability. The ridge regression probe is trained on non-perturbed sentences from the generic text corpus. For evaluation, we pool BLiMP, CoLA, and SyntaxGym into a single held-out dataset (see \S\ref{probe_prob}).

As shown in \hyperref[tab:ridge]{Table~\ref*{tab:ridge}}, ridge regression achieves moderately high $R^2$ when predicting prefix logprob from per-token hidden states and the last-toke hidden states. The gap between the two setups may partially be explained by the lower variance of the length-normalized prefix logprob at the final token versus across all tokens (\S ~\ref{logprob_var}), which can naturally limit the achievable $R^2$. The strong out-of-domain performance of the surprisal probe suggests that token-level hidden states retain substantial information about cumulative string probabilities.

\section{Related Work}
\paragraph{LMs as models of human linguistic cognition.}
Skeptics of LMs as models of human language learning argue that LMs statistically mimic  behaviors but do not possess the underlying structural rules of human cognition \citep{bender2021,katzir2023large, chomsky2023tyler, chomsky2023nyt}. Proponents, on the other hand, suggest that though LMs are not a complete theory of language, they nonetheless offer rich insights into the science of language \citep{portelancejasbi2024,futrell2025linguistics}. 

Central to the debate is inductive bias, i.e., priors brought by the learning systems beyond the training data distribution \citep{Mitchell1980,inductivebias}. Existing work has broadly investigated the inductive bias of LMs in learning syntactic principles, unnatural word orders, and (human) impossible languages \citep{fillergap, mitchell-bowers2020,kallini2024,xu2025}. In the same spirit, we treat LMs as a useful testbed for studying human-like linguistic behaviors and investigate the extent to which their representations encode syntactic knowledge.
\paragraph{LM acceptability judgments.}
Can language models trained with distributional objectives achieve human-level linguistic competence without strong linguistic inductive biases? One way to assess LM syntactic knowledge is metalinguistic acceptability judgment: given a sentence, models can be fine-tuned or few-shot prompted to determine if it is grammatically acceptable \cite{warstadt2019}. This approach is straightforward but assumes LMs faithfully follow instructions and have metalinguistic knowledge of what grammaticality means. Though metalinguistic prompting is not a major baseline in this work, we obtain and present the relevant results in \S\ref{fewshot_section}.

An alternative is to use LM string probability as a proxy for acceptability scores. \citet{Linzen2018} introduced minimal pairs—grammatical/ungrammatical sentences differing by small, targeted syntactic modifications—to test whether LMs assign higher probability to the grammatical variant, an approach now known as targeted syntactic evaluation \citep{warstadt2020blimp, jumelet2025}. However, probability-based acceptability scoring is sensitive to non-syntactic factors (e.g., input length, unigram frequency) and degrades outside pairwise comparisons \citep{sinha2023,tjuatja2025goeslmacceptabilityjudgment,hu-etal:2025-what-can-string-probability}. To address these limitations, our work shifts from using
string probability to computing acceptability scores from LM hidden states. 

\paragraph{Representations of syntactic information.}
One way to study the internal mechanisms of LMs is probing: training a low-capacity model (often linear or a small multilayer perceptron) to predict supervised-task features from an network's hidden states \citep{alain2018, Belinkov2022}. LMs have been shown to acquire nontrivial abstractions of syntactic dependencies, part-of-speech tags, parse trees, and incremental parse states, etc. \citep{lepori-mccoy-2020, belinkov2017, hewitt2019, eisape2022}. Though non-linear probes enable more flexible decoding of target features \citep{white2021, chen2021probing}, most prior work assumes that linguistic structures are linearly decodable from LM hidden states \citep{park2025}. Adopting this premise, we train linear probes to demonstrate that \emph{grammaticality}, a sentence-level syntactic property, is linearly extractable from LM representations.

\section{Conclusion}
We propose a simple approach to probe for representations of grammaticality in pretrained LMs. The trained probes selectively attend to grammaticality instead of semantic plausibility. We observe zero-shot cross-lingual transfer: a probe trained exclusively on English generalizes to a broad spectrum of typologically diverse languages. Finally, we find that that probe scores do not simply recapture probability. These results collectively suggest that despite being trained without built-in linguistic knowledge, LMs learn nontrivial representations of grammaticality that is distinct from output likelihoods. With the acknowledgment that LMs are fundamentally different systems from humans, our work invites further investigation into potential usages of LMs as models for linguistic processing and learning.

\section*{Limitations}
Our synthetic data creation is inherently limited. While we label the generic texts as grammatical, it is possible that they have built-in grammar errors. On the other hand, our synthetic perturbations may produce semantically implausible sentences rather than ungrammatical ones. Therefore, our training set are only pairs of \emph{mostly} grammatical/ungrammatical sentences.

Our work mainly concerns the notion of grammaticality in natural languages. It is likely that LMs also implicitly encode the syntactic acceptability in formal languages, but testing this hypothesis is outside the scope of this work. Any theory work in formal language learning is also beyond the empirical nature of this work.

Our work only examines the representation of LMs at their last checkpoints and does not include any learning-dynamics  results to study \emph{when} LMs develop the notion of grammaticality. Future work could investigate if grammaticality can be learned from human-plausible amount of training data.

\section*{Ethics Statement}
Although we draw analogies to human cognition (namely the competence–performance gap) and discuss the implications of LMs for linguistic cognition, we refrain from making claims about human language acquisition. This work restricts its scope to LMs as models of linguistic theories. Any conclusion of human linguistic processing derived from this work is unwarranted and strongly advised against

\section*{Acknowledgements}
This study was supported in part by the MIT-IBM Watson AI Lab. We thank the reviewers for the thorough reviews and constructive feedback. We thank Heidi Lei and Xiaoman Delores Ding for their constructive comments. Yingshan Susan Wang was funded by the MIT Undergraduate Research Opportunity Program. 

\bibliography{custom}

\appendix
\section{Contrastive Training Sets}\label{Perturb}
\paragraph{Generic text corpus.}
To build a generic English text corpus, we combine raw text from two sources:
(i) Penn Treebank (PTB; Wall Street Journal text-only splits) \citep{marcus1994}, and
(ii) the \href{https://huggingface.co/datasets/GenRM/gutenberg-dpo-v0.1-jondurbin}{\texttt{GenRM/gutenberg-dpo-v0.1-jondurbin}} dataset derived from Project Gutenberg books \citep{gutenbergdpo}.
For the Gutenberg Hugging Face dataset, we use only the \texttt{chosen} field, which contains the original (human-written)
book chapter texts.

\paragraph{Sentence extraction and sampling.}
We pool all strings (contiguous text blocks) from PTB and Gutenberg and shuffle the strings to interleave inputs from the two domains. We then segment each string into sentences using the PySBD segmenter. If a text block contains quoted spans, we extract the texts inside the quotes and re-segment them (discarding the non-quoted portions).
We keep the first 50{,}000 segmented sentences.

\paragraph{Vocabulary construction.} We construct a vocabulary for the \texttt{insertion} perturbation.
Given the 50{,}000 sentences, we extract tokens using RegEx to match contiguous alphabetic spans (\verb|[A-Za-z]+|) and map the found tokens to a Python set.

\paragraph{Insertion.} The first $\lfloor 50,000/3 \rfloor$ sentences are chosen for \texttt{insertion}. We tokenize each sentence with spaCy’s English tokenization. We then sample a number $k\in\{1,\dots,5\}$, choose $k$ random token boundaries in the sentence (including sentence beginning and end),
and insert $k$ vocabulary tokens sampled uniformly with replacement.
\paragraph{Deletion.}
The next $\lfloor 50{,}000/3 \rfloor$ sentences are assigned to \texttt{deletion} and tokenized with spaCy. For each sentence, we sample $k\in\{1,\dots,5\}$, delete $k$ uniformly chosen alphabetic tokens (\texttt{token.is\_alpha}) if possible and skip otherwise.

\paragraph{Local shuffle.}
The remaining sentences are assigned to \texttt{local shuffle} and tokenized with spaCy. For each sentence with at least 5 tokens, we select a random contiguous 5-token window and randomly permute those 5 tokens. If the sentences have fewer than 5 tokens, we skip them.

Finally, we return all the perturbed sentences with their original counterparts in pairs.

\paragraph{Validation of Synthetic Data.} 
We used Claude Opus 4.6 as an acceptability judge on a random sample of 5,000 original and 5,000 perturbed sentences from the synthetic dataset:                    
\begin{itemize}[leftmargin=*]
    \item Original ("good") sentences: 93.72\% judged acceptable.  
    \item Perturbed ("bad") sentences: 6.28\% judged acceptable. 
\end{itemize}
The LLM-as-judge results validate that our perturbation pipeline reliably geneates ungrammatical sentences. 
\section{Evaluation Datasets}
The specific statistics of syntactic acceptability evaluation datasets are listed in \hyperref[tab:syntactic_sets]{Table ~\ref{tab:syntactic_sets}}. All of them are publicly available on Hugging Face.

SyntaxGym is grouped by sentence conditions. We label conditions as acceptable vs. unacceptable using:
\begin{quote}\small
\textbf{Acceptable:}
np\_match, vp\_match, that\_nogap, what\_subjgap, what\_gap, neg\_pos, neg\_neg, match\_sing, match\_plural, no-sub\_no-matrix, sub\_matrix.\\
\textbf{Unacceptable:}
np\_mismatch, vp\_mismatch, what\_nogap, that\_subjgap, what\_matrixgap, that\_matrixgap, that\_gap, pos\_pos, pos\_neg, mismatch\_sing, mismatch\_plural, sub\_no-matrix, no-sub\_matrix.
\end{quote}

All semantic plausibility benchmarks come from \citet{eventknowledge}, and its three datasets are adapted from previous studies \citep{fedorenko2020, vassallo2018, Ivanova2021}. See \hyperref[tab:plausibility_sets]{Table ~\ref{tab:plausibility_sets}} for relevant information.
\label{sec:appendix}
\begin{table*}
    \centering
    \small
    \begin{tabular*}{\textwidth}{l@{\extracolsep{\fill}}lcccl}
        \toprule
        \textbf{Dataset} & \textbf{Language} & \textbf{\# Sentences} & \textbf{\% Grammatical} & \textbf{Minimal Pairs} & \textbf{References} \\
        \midrule
        BLiMP & English & 134,000 & 50.0 & \usym{2713} & \citet{warstadt2020blimp} \\
        CoLA & English & \phantom{0}10,657 & 70.4 & $\usym{2717}$ & \citet{warstadt2019} \\
        SyntaxGym & English & \phantom{00}2,412 & 49.0 & $\usym{2717}$ & \citet{gauthier2020} \\
        ScaLA (sv) & Swedish & \phantom{0}10,762 & 50.0 & $\usym{2717}$ & \citet{nielsen2023} \\
        BLiMP-NL & Dutch & \phantom{0}18,000 & 50.0 & \usym{2713} & \citet{suijkerbuijk2025} \\
        ItaCoLA & Italian & \phantom{00}8,776 & 84.4 & $\usym{2717}$ & \citet{Trotta2021} \\
        RuCoLA & Russian & \phantom{0}13,445 & 57.6 & $\usym{2717}$ & \citet{mikhailov2022} \\
        JCoLA & Japanese & \phantom{00}9,154 & 81.9 & $\usym{2717}$ & \citet{someya2024} \\
        SLING & Chinese & \phantom{0}80,000 & 50.0 & \usym{2713} & \citet{song2022} \\
        \bottomrule
    \end{tabular*}

    \caption{Overview of linguistic acceptability datasets for evaluation.}
    \label{tab:syntactic_sets}
\end{table*}

\section{Language Model Details}
All inference is run on local H100 and A100 clusters using hugginngface APIs. The model-specific statistics are given in \hyperref[tab:model_specs]{Table~\ref{tab:model_specs}}.
\begin{table*}[h]
\centering
\small
\begin{tabular}{l l c c c}
\toprule
\textbf{Model} & \textbf{Hugging Face ID} & \textbf{Hidden Dim} & \textbf{\# Layers} & \textbf{\# Training Tokens}  \\
\midrule
OLMo-2-7B & \texttt{allenai/OLMo-2-1124-7B} & 4096 & 32 & \phantom{0}4T \\
OLMo-3-7B & \texttt{allenai/Olmo-3-1025-7B} & 4096 & 32 & \phantom{0}6T \\
Llama-3.2-1B & \texttt{meta-llama/Llama-3.2-1B} & 2048 & 16 & \phantom{0}9T \\
Llama-3.1-8B & \texttt{meta-llama/Llama-3.1-8B} & 4096 & 32 & 15T \\
Gemma-2-2B & \texttt{google/gemma-2-2b} & 2304 & 26 & \phantom{0}2T \\
Gemma-2-9B & \texttt{google/gemma-2-9b} & 3584 & 42 & \phantom{0}8T \\
\bottomrule
\end{tabular}

\caption{Statistics of selected OLMo, Llama, and Gemma models.}
\end{table*}
\label{tab:model_specs}

\section{Probe Training Details}
All logistic probes are trained using SnapML's \texttt{LogisticRegression} (GPU backend), with \texttt{dual=False}, \texttt{max\_iter=1000}, \texttt{random\_state=0}, \texttt{fit\_intercept=True}. All ridge regressions are trained with SnapML's \texttt{LinearRegression} (GPU backend) with \texttt{fit\_intercept=True}. All probing experiments could be completed in fewer than 15 hours, with the LASSO probe’s neuron-selection process accounting for the majority (about 8 hours).
\paragraph{$\ell_2$ probe.}\label{l2}
\begin{figure}[h!] 
\includegraphics[width=\linewidth]{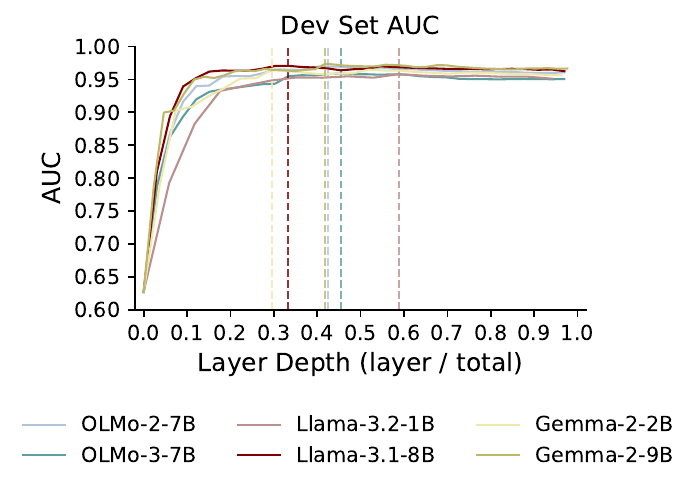}
  \caption{By-layer $\ell_2$-probe AUC on the dev set. Vertical lines indicate the best layers selected for evaluation (Llama-3.2-1B: 10, Llama-3.1-8B: 11, Gemma-2-2B: 8, Gemma-2-9B: 18, OLMo-2-7B: 14, OLMo-3-7B: 15).}
  \label{fig:by_layer}
\end{figure}

\begin{figure}[h!] 
\includegraphics[width=\linewidth]{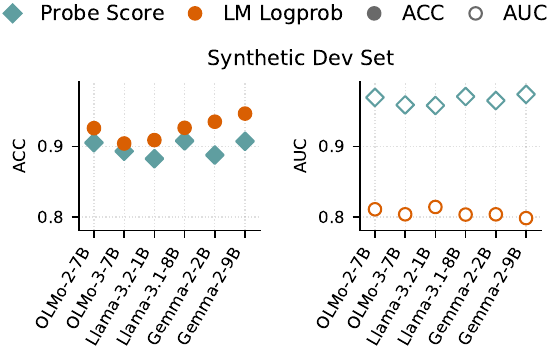}
  \caption{Comparing the performance of the best $\ell_2$-probes and LM Logprob on heldout in-domain dev set.}
  \label{fig:dev_final_metric}
\end{figure}

\begin{table*}[h!]
    \centering
    \small
\begin{tabular*}{\textwidth}{l@{\extracolsep{\fill}}lcccl}
        \toprule
        \textbf{Dataset} & \textbf{Language} & \textbf{\# Sentences} & \textbf{\% Plausible} & \textbf{Minimal Pairs} & \textbf{References} \\
        \midrule
        Dataset 1 & English & 1,564 & 50.0 & \usym{2713} & \citet{eventknowledge,fedorenko2020} \\
        Dataset 2 & English & \phantom{0,}790 & 50.0 & \usym{2713} & \citet{eventknowledge,vassallo2018} \\
        Dataset 3 & English & \phantom{0,0}76 & 50.0 & \usym{2713} & \citet{eventknowledge,Ivanova2021} \\
        \bottomrule
    \end{tabular*}
    \caption{Overview of semantic plausibility datasets for evaluation.}
    \label{tab:plausibility_sets}
\end{table*}

\begin{table*}[h!]
\centering
\small
\renewcommand{\arraystretch}{1.2}
\begin{tabular}{l l l p{8cm}}
\toprule
\textbf{Dataset} & \textbf{Structure} & \textbf{Condition / Label} & \textbf{Example Sentence} \\
\midrule
\multirow{2}{*}{\textbf{BLiMP}} & \multirow{2}{*}{\shortstack[l]{Minimal Pair\\(Matched)}} 
 & Acceptable & The cats licked themselves. \\
 & & Unacceptable & The cats licked itself. \\
\midrule
\multirow{2}{*}{\textbf{CoLA}} & \multirow{2}{*}{\shortstack[l]{Individual\\Sentences}} 
 & Acceptable & The book was written by John. \\
 & & Unacceptable & Books were sent to each other by the students. \\
\midrule
\multirow{4}{*}{\textbf{SyntaxGym}} & \multirow{4}{*}{\shortstack[l]{Test Suite\\(Grouped by Item)}} 
 & Cond. A & The farmer near the clerks knows many people. \\
 & & Cond. B & The farmer near the clerks know many people. \\
 & & Cond. C & The farmers near the clerk knows many people. \\
 & & Cond. D & The farmers near the clerk know many people. \\
\midrule
\multirow{2}{*}{\shortstack[l]{\textbf{Plausibility}\\\textbf{Dataset 1}}} & \multirow{2}{*}{\shortstack[l]{Minimal Pair\\(Matched)}} 
 & Plausible & The actor won the award. \\
 & & Implausible & The actor won the battle. \\
\bottomrule
\end{tabular}
\caption{Examples from different evaluation datasets.}
\label{tab:dataset-examples}
\end{table*}
For each layer, we train an $\ell_2$-regularized logistic regression probe on the last-token hidden-state vector, using an 80/20 train--dev split with feature normalization. We preprocess the hidden states by computing per-dimension mean and standard deviation on the training split and z-scoring both train and dev features with these statistics. We sweep the regularization strength over a log-scale grid, $\alpha \in \{2^{s}, 2^{s+1}, \ldots, 2^{e}\}$ for $s=-2$, $e=5$, and select the best hyperparameter per layer by maximizing validation AUC. We choose the layer whose probe achieves the highest AUC on the dev set (\hyperref[fig:by_layer]{Figure~\ref{fig:by_layer}}).

\paragraph{LASSO probe.}\label{l1}
We train an $\ell_1$-regularized logistic regression probe (LASSO) on concatenated last-token hidden states of all model layers. We tune $\ell_1$ penalty strength to achieve a target sparsity level. For each target fraction $p \in$ \{0.01\%, 0.05\%, 0.1\%, 0.5\%\}, we set the desired number of active neurons to $k=\lceil pD\rceil$ where $D$ is the total number of neurons and adaptively adjust the $\ell_1$ penalty strength until the fitted probe has $k'$ non-zero coefficients for $|k - k'| \leq 0.05k$. We then treat the non-zero dimensions as the selected neuron set and refit an $\ell_2$-regularized logistic regression probe restricted to these neurons. For this refit, we sweep $\alpha \in \{2^{s}, 2^{s+1}, \ldots, 2^{e}\}$ with $s=-2$ and $e=5$, and select the best $\alpha$ by validation AUC. As a baseline, we repeat the $\ell_2$ refit on 30 random subsets of $k'$ neurons and report the average performance.

\paragraph{Probes augmented with probabilities.}\label{probe_plus_prob}
For each sentence $x$, we compute the LM's  length normalized string probabilities $\tilde{\ell}(x)$.
We then concatenate this scalar to the representation vector used by the probe, yielding an augmented feature vector
$z \;=\; [\,h \,;\, \tilde{\ell}\,],$ where $h$ denotes the best-layer hidden-state features chosen in \S\ref{l2}. We train the linear probe exactly as in \S\ref{l2}, treating the probability feature as an additional input dimension.

\paragraph{Probing for probabilities.}\label{probe_prob}
We train a ridge regression probe to predict the model’s token-level probability from hidden states. Concretely, from the forward pass of the \emph{original sentences} in the generic text corpus, we collect pairs $(h_t, \tilde{\ell}_t)$, where $h_t$ is the the best layer hidden states and $\tilde{\ell}_t$ is the length-normalized cumulative logprob at time $t$. We z-score $h_t$ using per-dimension mean and standard deviation computed on the training split. We use an 80/20 train--dev split and sweep the $\ell_2$ penalty over $\alpha \in \{2^{s}, 2^{s+1}, \ldots, 2^{e}\}$ with $s=-2$, $e=5$, selecting the best $\alpha$ by minimizing dev MSE. In addition, we train a last-token-only variant of the probe by restricting the input features and labels to time $t=T$ for each sentence.
\begin{figure}[h!] 
\section{Performance of Supervised Probes}\label{supervised_probe}
\includegraphics[width=\linewidth]{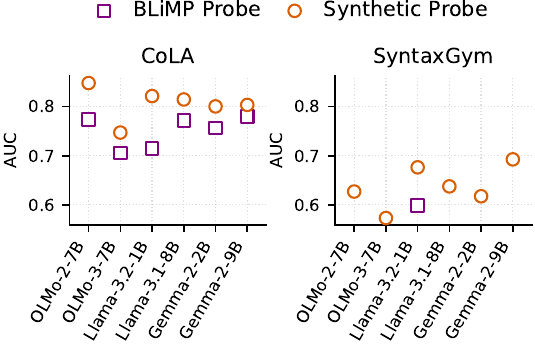}
  \caption{Comparisons of unsupervised (trained on synthetic data) and supervised probes (trained on BLiMP).}
\end{figure}
\begin{table}[h]
\centering
\small
\begin{tabular}{lcc}
    \toprule
    & \multicolumn{2}{c}{\textbf{BLiMP Probe $\rightarrow$ Synthetic Set}} \\
    \cmidrule(lr){2-3}
    \textbf{Models} & \textbf{AUC} & \textbf{ACC}\\
    \midrule
    Llama-3.2-1B & 0.97 & 0.99 \\
    Llama-3.1-8B & 0.98 & 0.99 \\
    \midrule
    Gemma-2-2B   & 0.97 & 0.99 \\
    Gemma-2-9B   & 0.98 & 0.99 \\
    \midrule
    OLMo-2-7B    & 0.98 & 0.99 \\
    OLMo-3-7B    & 0.97 & 0.99 \\
    \bottomrule
\end{tabular}
    \caption{Probe trained on BLiMP achieves near perfect performance on our synthetic data.}
    \label{tab:blimptosynthetic}
\end{table}

\noindent To establish a supervised baseline, we utilize the same selected layers (\hyperref[fig:by_layer]{Figure ~\ref{fig:by_layer}}) to train probes on BLiMP, subsequently evaluating them on CoLA, SyntaxGym, and our synthetic data.

\begin{table*}[t]
\section{$\ell_2$ Probe Performance on BLiMP by Linguistic Term}
\centering
\resizebox{\textwidth}{!}{%
\begin{tabular}{llcccccccccccc}
\toprule
\multirow{2}{*}{\textbf{Linguistic Term}} & \multirow{2}{*}{\textbf{Method}} & \multicolumn{2}{c}{\textbf{OLMo-2-7B}} & \multicolumn{2}{c}{\textbf{OLMo-3-7B}} & \multicolumn{2}{c}{\textbf{Llama-3.2-1B}} & \multicolumn{2}{c}{\textbf{Llama-3.1-8B}} & \multicolumn{2}{c}{\textbf{Gemma-2-2B}} & \multicolumn{2}{c}{\textbf{Gemma-2-9B}} \\
\cmidrule(lr){3-4} \cmidrule(lr){5-6} \cmidrule(lr){7-8} \cmidrule(lr){9-10} \cmidrule(lr){11-12} \cmidrule(lr){13-14}
 & & \textbf{AUC} & \textbf{ACC} & \textbf{AUC} & \textbf{ACC} & \textbf{AUC} & \textbf{ACC} & \textbf{AUC} & \textbf{ACC} & \textbf{AUC} & \textbf{ACC} & \textbf{AUC} & \textbf{ACC} \\
\midrule
\multirow{2}{*}{Anaphor Agreement} & LM Logprob & 0.69 & 0.99 & 0.69 & 0.99 & 0.68 & 0.99 & 0.69 & 0.98 & 0.67 & 0.99 & 0.65 & 0.99 \\
 & Probe Score & 0.85 & 0.96 & 0.81 & 0.90 & 0.80 & 0.95 & 0.89 & 0.96 & 0.78 & 0.93 & 0.93 & 0.98 \\
\midrule
\multirow{2}{*}{Argument Structure} & LM Logprob & 0.60 & 0.70 & 0.62 & 0.73 & 0.62 & 0.73 & 0.61 & 0.72 & 0.57 & 0.66 & 0.57 & 0.65 \\
 & Probe Score & 0.76 & 0.81 & 0.74 & 0.79 & 0.80 & 0.85 & 0.78 & 0.82 & 0.80 & 0.84 & 0.80 & 0.85 \\
\midrule
\multirow{2}{*}{Binding} & LM Logprob & 0.62 & 0.81 & 0.61 & 0.78 & 0.61 & 0.78 & 0.62 & 0.81 & 0.58 & 0.78 & 0.58 & 0.80 \\
 & Probe Score & 0.71 & 0.82 & 0.65 & 0.75 & 0.67 & 0.80 & 0.73 & 0.84 & 0.69 & 0.79 & 0.77 & 0.87 \\
\midrule
\multirow{2}{*}{Control / Raising} & LM Logprob & 0.64 & 0.79 & 0.65 & 0.81 & 0.65 & 0.81 & 0.64 & 0.79 & 0.61 & 0.76 & 0.60 & 0.74 \\
 & Probe Score & 0.72 & 0.83 & 0.70 & 0.77 & 0.72 & 0.83 & 0.71 & 0.80 & 0.77 & 0.88 & 0.76 & 0.86 \\
\midrule
\multirow{2}{*}{\shortstack[l]{Determiner-Noun\\Agreement}} & LM Logprob & 0.67 & 0.91 & 0.67 & 0.93 & 0.67 & 0.92 & 0.66 & 0.91 & 0.64 & 0.93 & 0.63 & 0.93 \\
 & Probe Score & 0.91 & 0.97 & 0.81 & 0.90 & 0.91 & 0.96 & 0.94 & 0.98 & 0.97 & 0.99 & 0.96 & 0.99 \\
\midrule
\multirow{2}{*}{Ellipsis} & LM Logprob & 0.63 & 0.84 & 0.63 & 0.87 & 0.63 & 0.89 & 0.63 & 0.86 & 0.59 & 0.84 & 0.59 & 0.82 \\
 & Probe Score & 0.69 & 0.78 & 0.63 & 0.71 & 0.65 & 0.77 & 0.69 & 0.78 & 0.64 & 0.71 & 0.67 & 0.74 \\
\midrule
\multirow{2}{*}{Filler Gap} & LM Logprob & 0.58 & 0.72 & 0.60 & 0.77 & 0.59 & 0.78 & 0.57 & 0.73 & 0.54 & 0.69 & 0.54 & 0.67 \\
 & Probe Score & 0.71 & 0.81 & 0.68 & 0.76 & 0.67 & 0.76 & 0.74 & 0.83 & 0.75 & 0.87 & 0.73 & 0.83 \\
\midrule
\multirow{2}{*}{Irregular Forms} & LM Logprob & 0.68 & 0.90 & 0.73 & 0.96 & 0.70 & 0.92 & 0.70 & 0.92 & 0.66 & 0.90 & 0.71 & 0.96 \\
 & Probe Score & 0.89 & 0.98 & 0.85 & 0.94 & 0.87 & 0.98 & 0.88 & 0.97 & 0.91 & 1.00 & 0.92 & 0.97 \\
\midrule
\multirow{2}{*}{Island Effects} & LM Logprob & 0.61 & 0.72 & 0.62 & 0.72 & 0.59 & 0.68 & 0.59 & 0.70 & 0.55 & 0.64 & 0.55 & 0.65 \\
 & Probe Score & 0.73 & 0.80 & 0.70 & 0.74 & 0.63 & 0.70 & 0.71 & 0.74 & 0.71 & 0.79 & 0.72 & 0.78 \\
\midrule
\multirow{2}{*}{NPI Licensing} & LM Logprob & 0.58 & 0.71 & 0.58 & 0.72 & 0.58 & 0.71 & 0.56 & 0.68 & 0.57 & 0.67 & 0.59 & 0.71 \\
 & Probe Score & 0.79 & 0.91 & 0.77 & 0.88 & 0.75 & 0.87 & 0.70 & 0.80 & 0.78 & 0.90 & 0.81 & 0.91 \\
\midrule
\multirow{2}{*}{Quantifiers} & LM Logprob & 0.63 & 0.75 & 0.58 & 0.64 & 0.64 & 0.76 & 0.63 & 0.75 & 0.57 & 0.66 & 0.56 & 0.62 \\
 & Probe Score & 0.64 & 0.72 & 0.63 & 0.70 & 0.70 & 0.80 & 0.64 & 0.75 & 0.63 & 0.71 & 0.68 & 0.84 \\
\midrule
\multirow{2}{*}{\shortstack[l]{Subject-Verb\\Agreement}} & LM Logprob & 0.60 & 0.79 & 0.64 & 0.88 & 0.61 & 0.82 & 0.61 & 0.82 & 0.60 & 0.83 & 0.61 & 0.83 \\
 & Probe Score & 0.77 & 0.91 & 0.76 & 0.88 & 0.82 & 0.95 & 0.84 & 0.94 & 0.85 & 0.97 & 0.86 & 0.95 \\
\bottomrule
\end{tabular}%
}
\caption{BLiMP results broken down by different categories. We report the performance of best-layer $\ell_2$ probe scores and LM logprob.}
\label{tab:blimp_results}
\end{table*}

\begin{table*}[t]
\section{$\ell_2$-Probe Confidence Intervals}\label{CI}
\centering
\small
\setlength{\tabcolsep}{10pt} %
\begin{tabular}{llcccc}
\toprule
\multirow{2}{*}{\textbf{Models}} & \multirow{2}{*}{\textbf{Methods}} & \multicolumn{2}{c}{\textbf{BLiMP}} & \textbf{CoLA} & \textbf{SyntaxGym} \\
\cmidrule(lr){3-4} \cmidrule(lr){5-5} \cmidrule(lr){6-6}
 &  & \textbf{AUC} & \textbf{ACC} & \textbf{AUC} & \textbf{AUC} \\
\midrule
\multirow{2}{*}{Llama-3.2-1B} 
 & LM Logprob  & [0.61, 0.61] & [0.79, 0.79] & [0.67, 0.69] & [0.50, 0.55] \\
 & Probe Score & [0.73, 0.74] & [0.84, 0.84] & [0.78, 0.80] & [0.66, 0.70] \\
\midrule
\multirow{2}{*}{Llama-3.1-8B} 
 & LM Logprob  & [0.60, 0.61] & [0.78, 0.78] & [0.68, 0.70] & [0.51, 0.55] \\
 & Probe Score & [0.76, 0.76] & [0.84, 0.85] & [0.81, 0.82] & [0.61, 0.66] \\
\midrule
\multirow{2}{*}{Gemma-2-2B}   
 & LM Logprob  & [0.58, 0.58] & [0.75, 0.75] & [0.62, 0.65] & [0.49, 0.54] \\
 & Probe Score & [0.77, 0.77] & [0.86, 0.87] & [0.79, 0.81] & [0.59, 0.64] \\
\midrule
\multirow{2}{*}{Gemma-2-9B}   
 & LM Logprob  & [0.58, 0.58] & [0.75, 0.75] & [0.62, 0.65] & [0.49, 0.54] \\
 & Probe Score & [0.79, 0.79] & [0.88, 0.88] & [0.79, 0.81] & [0.67, 0.71] \\
\midrule
\multirow{2}{*}{OLMo-2-7B}    
 & LM Logprob  & [0.61, 0.61] & [0.77, 0.78] & [0.67, 0.69] & [0.51, 0.55] \\
 & Probe Score & [0.75, 0.76] & [0.85, 0.85] & [0.78, 0.80] & [0.60, 0.65] \\
\midrule
\multirow{2}{*}{OLMo-3-7B}    
 & LM Logprob  & [0.61, 0.62] & [0.79, 0.80] & [0.68, 0.70] & [0.52, 0.56] \\
 & Probe Score & [0.71, 0.72] & [0.80, 0.81] & [0.74, 0.76] & [0.55, 0.60] \\
\bottomrule
\end{tabular}
\caption{95\% confidence intervals of $\ell_2$-probe performance metrics (AUC/ACC) for BLiMP, CoLA, and SyntaxGym.}
\label{tab:main_results}
\end{table*}

\begin{table*}[t]
\centering
\small
\setlength{\tabcolsep}{6pt}
\begin{tabular}{llcccc}
\toprule
\multirow{2}{*}{\textbf{Models}} & \multirow{2}{*}{\textbf{Methods}} & \textbf{ScaLA (sr)} & \multicolumn{2}{c}{\textbf{BLiMP-NL}} & \textbf{ItaCoLA} \\
\cmidrule(lr){3-3} \cmidrule(lr){4-5} \cmidrule(lr){6-6} 
 &  & \textbf{AUC} & \textbf{AUC} & \textbf{ACC} & \textbf{AUC} \\
\midrule
\multirow{2}{*}{Llama-3.2-1B} 
  & LM Logprob  & [0.61, 0.63] & [0.57, 0.59] & [0.74, 0.76] & [0.56, 0.59] \\
  & Probe Score & [0.67, 0.69] & [0.58, 0.60] & [0.68, 0.70] & [0.57, 0.61] \\
\midrule
\multirow{2}{*}{Llama-3.1-8B} 
  & LM Logprob  & [0.64, 0.66] & [0.61, 0.63] & [0.84, 0.85] & [0.60, 0.63] \\
  & Probe Score & [0.73, 0.75] & [0.65, 0.66] & [0.76, 0.77] & [0.64, 0.67] \\
\midrule
\multirow{2}{*}{Gemma-2-2B}   
  & LM Logprob  & [0.63, 0.65] & [0.60, 0.62] & [0.81, 0.83] & [0.53, 0.57] \\
  & Probe Score & [0.68, 0.70] & [0.62, 0.63] & [0.72, 0.74] & [0.60, 0.63] \\
\midrule
\multirow{2}{*}{Gemma-2-9B}   
  & LM Logprob  & [0.65, 0.67] & [0.62, 0.63] & [0.85, 0.87] & [0.54, 0.58] \\
  & Probe Score & [0.82, 0.84] & [0.73, 0.74] & [0.83, 0.85] & [0.69, 0.72] \\
\bottomrule
\end{tabular}
\label{tab:results_part1}
\caption{95\% confidence intervals of $\ell_2$-probe performance metrics (AUC/ACC) for ScaLA (sv) (Swedish), BLiMP-NL (Dutch), and ItaCoLA (Italian).}
\end{table*}

\begin{table*}[t]
\centering
\small
\setlength{\tabcolsep}{6pt}
\begin{tabular}{llcccc}
\toprule
\multirow{2}{*}{\textbf{Models}} & \multirow{2}{*}{\textbf{Methods}} & \textbf{RuCoLA} & \textbf{JCoLA} & \multicolumn{2}{c}{\textbf{SLiNG}} \\
\cmidrule(lr){3-3} \cmidrule(lr){4-4} \cmidrule(lr){5-6}
 &  & \textbf{AUC} & \textbf{AUC} & \textbf{AUC} & \textbf{ACC} \\
\midrule
\multirow{2}{*}{Llama-3.2-1B} 
  & LM Logprob  & [0.43, 0.45] & [0.55, 0.58] & [0.55, 0.56] & [0.58, 0.59] \\
  & Probe Score & [0.58, 0.60] & [0.55, 0.59] & [0.57, 0.57] & [0.62, 0.63] \\
\midrule
\multirow{2}{*}{Llama-3.1-8B} 
  & LM Logprob  & [0.45, 0.47] & [0.57, 0.61] & [0.56, 0.57] & [0.65, 0.66] \\
  & Probe Score & [0.56, 0.58] & [0.61, 0.64] & [0.60, 0.60] & [0.68, 0.69] \\
\midrule
\multirow{2}{*}{Gemma-2-2B}   
  & LM Logprob  & [0.45, 0.47] & [0.58, 0.61] & [0.57, 0.58] & [0.61, 0.62] \\
  & Probe Score & [0.60, 0.62] & [0.59, 0.62] & [0.63, 0.64] & [0.73, 0.74] \\
\midrule
\multirow{2}{*}{Gemma-2-9B}   
  & LM Logprob  & [0.46, 0.48] & [0.58, 0.61] & [0.57, 0.58] & [0.61, 0.62] \\
  & Probe Score & [0.64, 0.66] & [0.68, 0.71] & [0.67, 0.67] & [0.75, 0.76] \\
\bottomrule
\end{tabular}
\caption{95\% confidence intervals of $\ell_2$-probe performance metrics for RuCoLA (Russian), JCoLA (Japanese), and SLiNG (Chinese).}
\label{tab:results_part2}
\end{table*}

\begin{figure*}[h!] 
\section{By-Layer Distributions of LASSO-Selected Neurons}
\includegraphics[width=\linewidth]{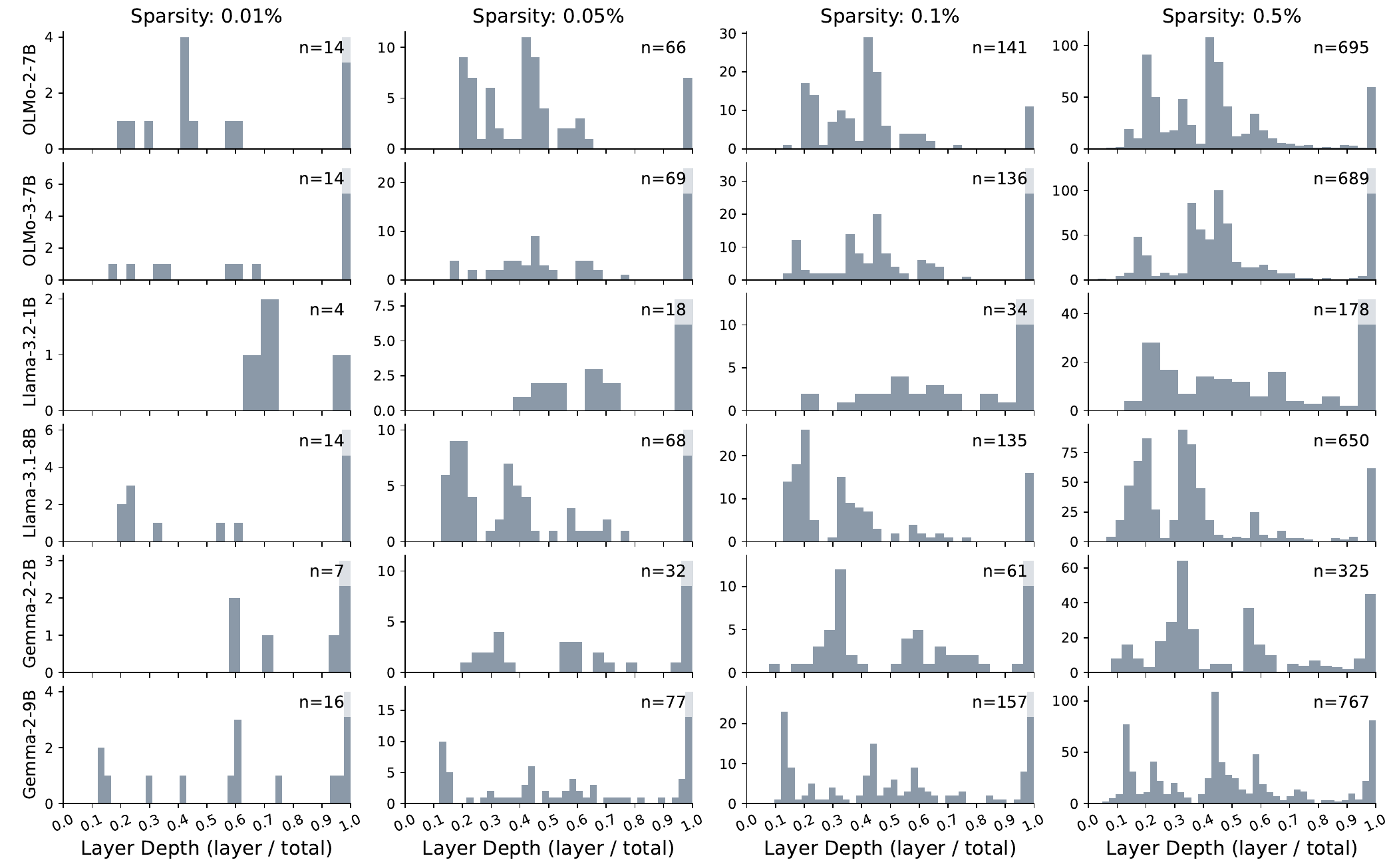}
  \caption{By-layer distributions of LASSO-selected neurons. The important neurons are distributed across many layers, with high concentration in the last layer.}
  \label{fig:l1_hist}
\end{figure*}

\begin{figure*}[t!]
    \section{More Distributions of LM Logprob and $\ell_2$ Probe Scores}
    \vspace{0.5em} %
    
    \begin{minipage}[t]{0.49\textwidth}
        \centering
        \includegraphics[width=\linewidth]{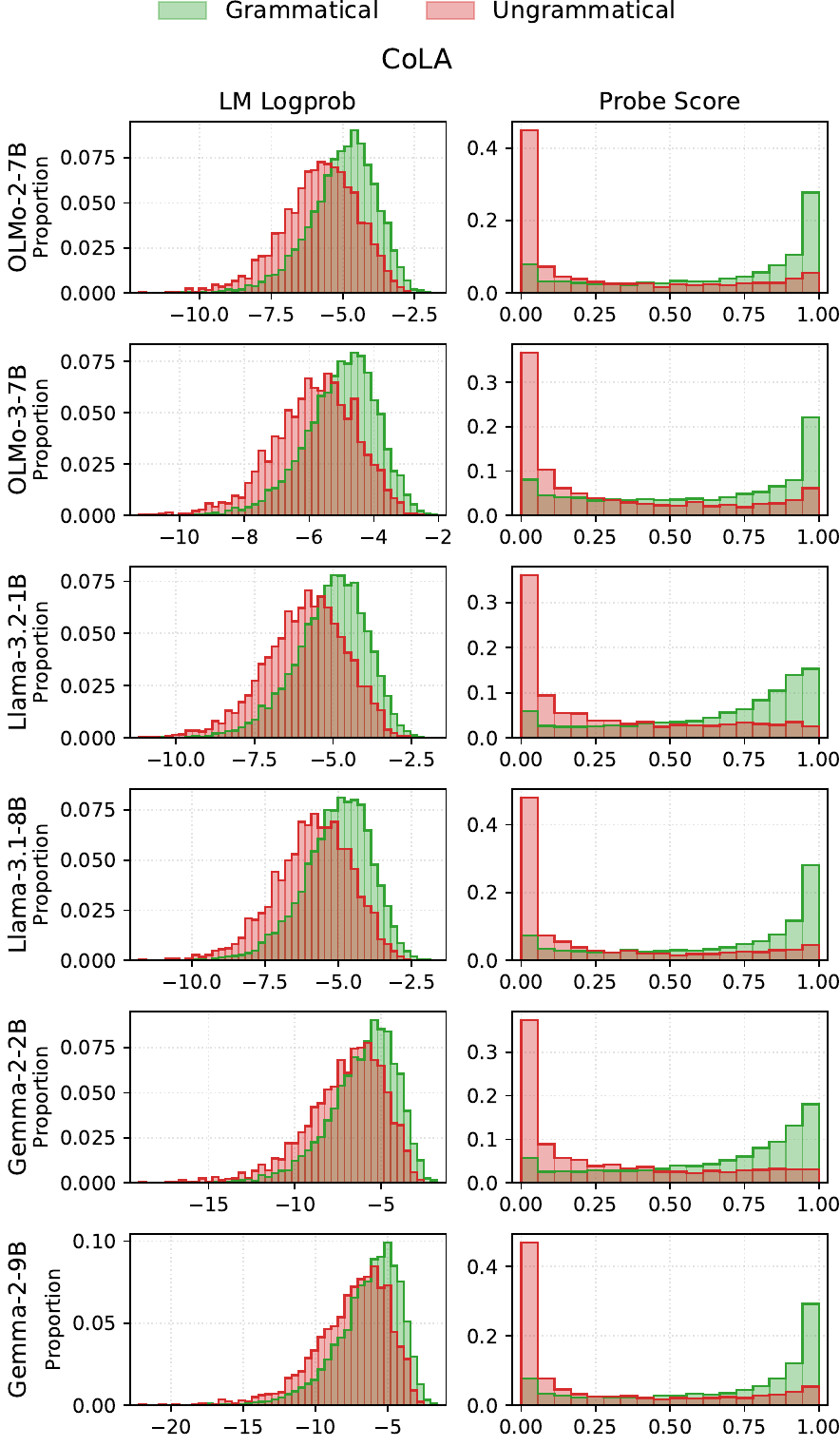}
        \caption{Distributions of LM logprob and probe scores on CoLA.}
        \label{fig:dist_cola}
    \end{minipage}
    \hfill 
    \begin{minipage}[t]{0.49\textwidth}
        \centering
        \includegraphics[width=\linewidth]{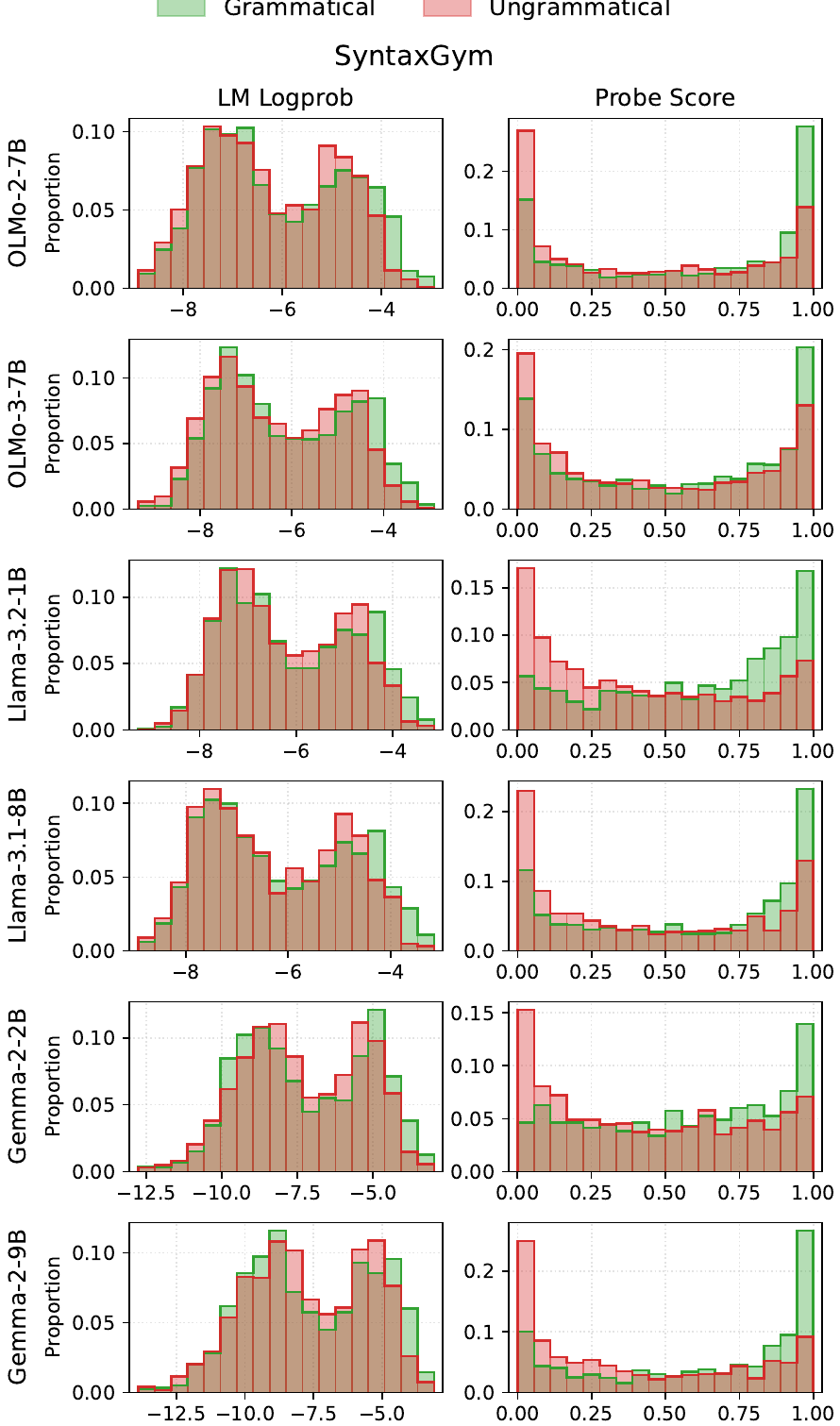} 
        \caption{Distributions of LM logprob and probe scores on SyntaxGym.}
        \label{fig:dist_syngym}
    \end{minipage}
\end{figure*}

\begin{figure}[h!]
    \includegraphics[width=\linewidth]{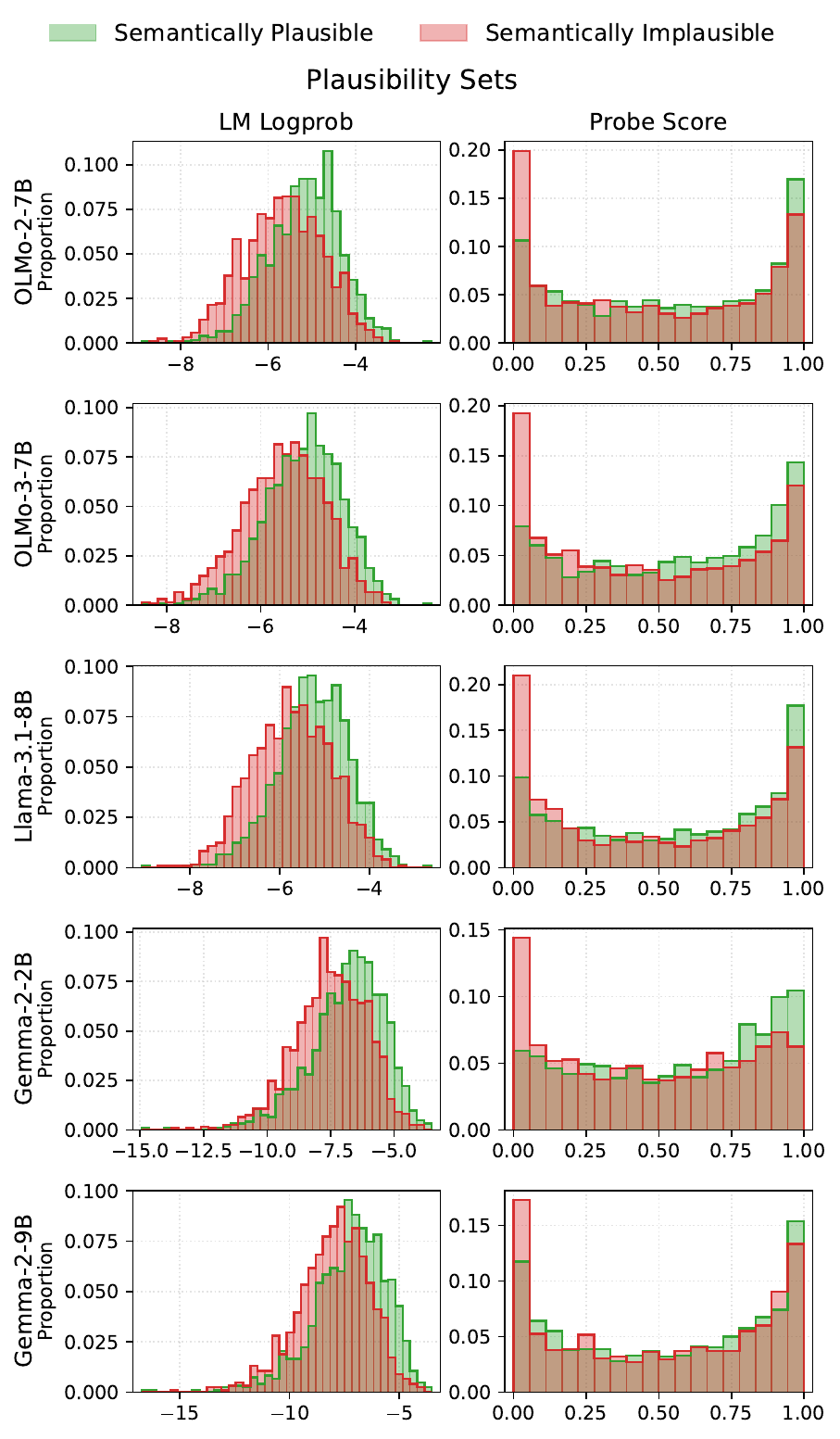} 
    \caption{Distributions of LM logprob and probe scores on Plausibility Sets.}
    \label{fig:dist_plausibility}
\end{figure}

\begin{figure}[h!] 
\section{LASSO Probe Performance on All Acceptability Datasets}
\includegraphics[width=\linewidth]{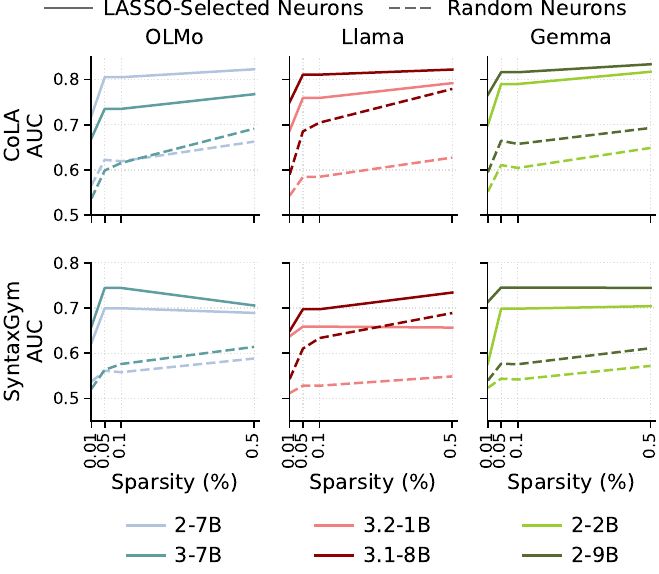}
  \caption{LASSO results on CoLA and SyntaxGym.}
  \label{fig:other_english_l1}
\end{figure}
\begin{figure}[h!] 
\includegraphics[width=\linewidth]{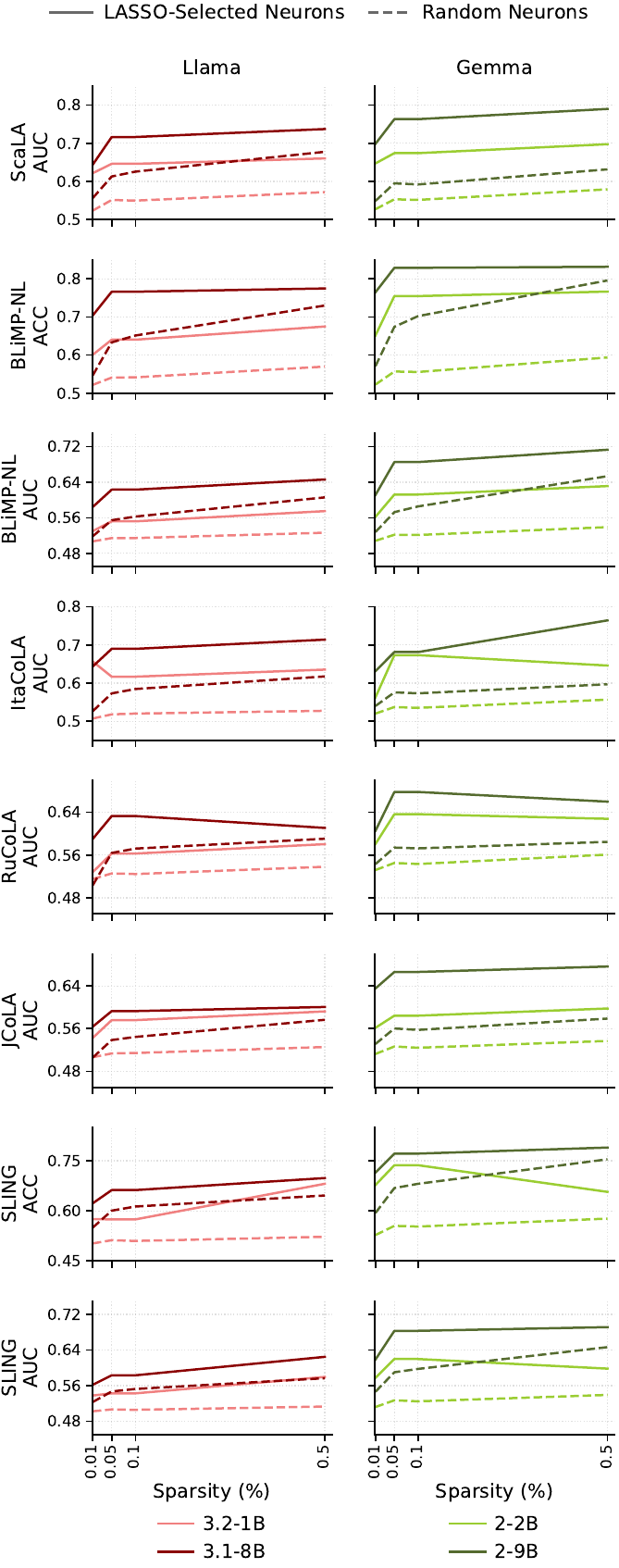}
  \caption{LASSO probe results on multilingual acceptability datasets for Llama and Gemma models.}
  \label{fig:multilingual_l1}
\end{figure}
\clearpage

\begin{table}[tbp]
\section{Pearson's Correlation of Length Normalized Logprob}
    \centering
    \small

    \begin{tabular}{lccc}
        \toprule
        & \multicolumn{3}{c}{\textbf{Corr.\,(logprob, log probe score)}} \\
        \cmidrule(lr){2-4}
        \textbf{Models} & \textbf{BLiMP} & \textbf{CoLA} & \textbf{SyntaxGym} \\
        \midrule
        Llama-3.2-1B & 0.24 & 0.34 & 0.07 \\
        Llama-3.1-8B & 0.22 & 0.36 & 0.23 \\
        \midrule
        Gemma-2-2B   & 0.22 & 0.29 & 0.18 \\
        Gemma-2-9B   & 0.18 & 0.29 & 0.26 \\
        \midrule
        OLMo-2-7B    & 0.19 & 0.30 & 0.26 \\
        OLMo-3-7B    & 0.25 & 0.40 & 0.21 \\
        \bottomrule
    \end{tabular}
    \caption{Pearson correlation coefficient of LM logprob and log probe scores. \textbf{Correlation strengths are moderate, suggesting that probe scores are not simply recapturing the LM logprob.}}
    \label{tab:PearsonR}
\end{table}

\begin{table}[h!]
\section{Variance of Length Normalized Cumulative Logprob}\label{logprob_var}
    \centering
    \small    
    \begin{tabular}{lccc}
        \toprule
        & \multicolumn{2}{c}{\textbf{Train Set Logprob Variance ($\sigma^2$)}} \\
        \cmidrule(lr){2-3}
        \textbf{Models} & \textbf{Per Token} & \textbf{Last Tok Only}\\
        \midrule
        Llama-3.2-1B & 1.99 & 1.32 \\
        Llama-3.1-8B & 2.32  & 1.45 \\
        \midrule
        Gemma-2-2B   & 5.12 & 2.22 \\
        Gemma-2-9B   & 6.64  & 2.75 \\
        \midrule
        OLMo-2-7B    & 2.18  & 1.49 \\
        OLMo-3-7B    & 2.23  & 1.36 \\
        \bottomrule
    \end{tabular}
    \caption{Logprob variance of original (assumed grammatical) sentences in the synthetic training corpus. Values are provided for the per-token and last-token-only configurations used in the ridge regression probes.\\}
    \begin{tabular}{lccc}
        \toprule
        & \multicolumn{2}{c}{\textbf{Eval Set Logprob Variance ($\sigma^2$)}} \\
        \cmidrule(lr){2-3}
        \textbf{Models} & \textbf{Per Token} & \textbf{Last Tok Only}\\
        \midrule
        Llama-3.2-1B & 2.83 & 1.32 \\
        Llama-3.1-8B & 2.43 & 1.29 \\
        \midrule
        Gemma-2-2B   & 11.97 & 4.15 \\
        Gemma-2-9B   &  16.41 & 5.46 \\
        \midrule
        OLMo-2-7B    & 2.65 & 1.35 \\
        OLMo-3-7B    & 1.33 & 2.77 \\
        \bottomrule
    \end{tabular}
    \caption{Logprob variance of sentences in the pooled evaluation sets (BLiMP, CoLA, and SyntaxGym). Values are provided for the per-token and last-token-only setups.}
\end{table}

\begin{table*}[t]                \section{Cross-lingual Transfer of Non-English Trained Probes for Llama-3.2-1B}\label{non_english_transfer}
  \centering                                  
  \small                                      
  \setlength{\tabcolsep}{5pt}                 
  \renewcommand{\arraystretch}{1.15}                      
  \begin{tabular}{@{}llcccccc@{}}
  \toprule
  & & \multicolumn{6}{c}{\textbf{Test Dataset (AUC)}} \\
  \cmidrule(lr){3-8}
  & & \textbf{Swedish} & \textbf{Dutch} & \textbf{Italian} & \textbf{Russian} & \textbf{Japanese} & \textbf{Chinese} \\
  \cmidrule(lr){3-3} \cmidrule(lr){4-4} \cmidrule(lr){5-5} \cmidrule(lr){6-6} \cmidrule(lr){7-7} \cmidrule(lr){8-8}
  \textbf{Language} & \textbf{Train Dataset}
    & \textbf{ScaLA (sv)} & \textbf{BLiMP-NL} & \textbf{ItaCoLA} & \textbf{RuCoLA} & \textbf{JCoLA} & \textbf{SLING} \\
  \midrule
  Swedish  & ScaLA (sv) & ---  & 0.58 & 0.59 & 0.53 & 0.52 & 0.56 \\
  Dutch    & BLiMP-NL   & 0.60 & ---  & 0.53 & 0.50 & 0.50 & 0.55 \\
  Italian  & ItaCoLA    & 0.58 & 0.53 & ---  & 0.51 & 0.53 & 0.49 \\
  Russian  & RuCoLA     & 0.57 & 0.54 & 0.53 & ---  & 0.52 & 0.49 \\
  Japanese & JCoLA      & 0.54 & 0.54 & 0.51 & 0.56 & ---  & 0.54 \\
  Chinese  & SLING      & 0.54 & 0.53 & 0.49 & 0.52 & 0.52 & ---  \\
  \bottomrule
  \end{tabular}

  \caption{Cross-lingual transfer AUC for Llama-3.2-1B. Each row indicates the training language and each column indicates the test
  language. Transfer from non-English languages is less effective than from English (\hyperref[tab:multilingual]{Table~\ref*{tab:multilingual}}), consistent with the hypothesized English-dominant representation space of the model.}
  \label{tab:crosslingual_transfer}
  \end{table*}

\section{Syntactic Acceptability Judgments by Metalinguistic Prompting}\label{fewshot_section}
To compare best-layer $\ell_2$ probes against the metalinguistic syntactic judgments, we employ few-shot prompting with the template shown in \hyperlink{fig:fewshot_template}{Figure~\ref{fig:fewshot_template}}. For a dataset of $N$ sentences, 
Let $\hat{y}_i \in \{0, 1\}$ denote the binary prediction for sentence $x_i$ generated either via the metalinguistic prompt or the $\ell_2$ probe (setting 0.5 as the probe score threshold). The non-pairwise accuracy is $
    \text{Accuracy} = \frac{1}{N} \sum_{i=1}^{N} \mathbbm{1}(\hat{y}_i = y_i),$
where $y_i \in \{0, 1\}$ is the ground-truth binary label. 

\begin{figure*}[h!] 
  
  \centering
  \includegraphics[width=\linewidth]{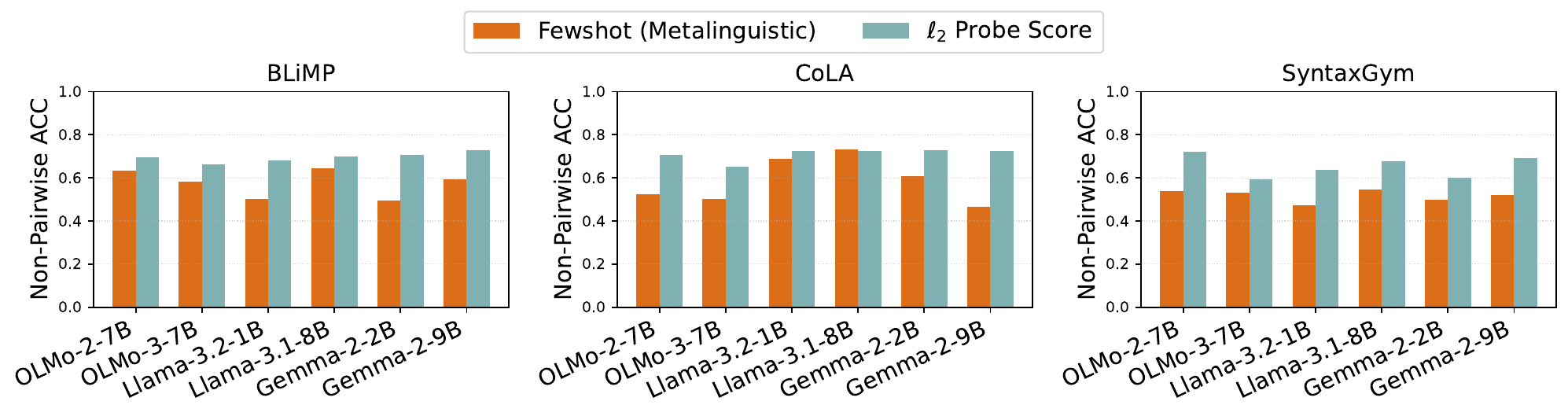}
  
  \caption{Nonpairwise accuracies of metalinguistic fewshot prompting and $\ell_2$ probes. The random baseline is 0.5.}
  \label{fig:fewshot}

\end{figure*}

As seen in \hyperlink{fig:fewshot}{Figure~\ref{fig:fewshot}}, our probes surpass the performance of metalinguistic judgments across all datasets for all models. Note that smaller models (Llama-3.2-1B and Gemma-2-2B) may not have much capability to learn in context by fewshot examples, which could explain their near random metalinguistic judgment performance on BLiMP.

\begin{figure*}[h!]
\centering
\begin{tcolorbox}[
    colback=templategray,
    colframe=gray,
    title=\textbf{Prompt Template for Metalinguistic Grammaticality judgments},
    fonttitle=\sffamily,
    boxrule=0.5pt,
    arc=2pt
]
\small
\ttfamily
Determine if the following sentences are grammatical.

\vspace{0.5em}

Sentence: The boy kick the ball. \\
Grammatical: No.

\vspace{0.5em}
Sentence: That you are back surprised me. \\
Grammatical: Yes.

\vspace{0.5em}
Sentence: The story goes on and on. \\
Grammatical: Yes.
\vspace{0.5em}

Sentence: Last night I was ever drunk. \\
Grammatical: No.
\vspace{0.5em}
\\
Sentence: \textcolor{keywordblue}{\{Input\_Sentence\}} \\
Grammatical: \textcolor{red}{\{Prediction\}}
\end{tcolorbox}
\caption{The metalinguistic prompting template. The model correctly answers the prompt if it outputs the corresponding grammaticality prediction of the input sentence (``Yes.'' or ``No.''). A prediction is classified as ``Yes.'' if its conditional probability exceeds that of ``No.''; otherwise, it is classified as ``No.''}
\label{fig:fewshot_template}
\end{figure*}

\end{document}